\documentclass[11pt]{article}
\pdfoutput=1

\PassOptionsToPackage{table}{xcolor}

\usepackage{acl}
\usepackage[T1]{fontenc}
\usepackage[utf8]{inputenc}
\usepackage{times}
\usepackage{latexsym}
\usepackage{microtype}
\usepackage{needspace}
\usepackage{inconsolata}
\usepackage{booktabs}
\usepackage{array}
\usepackage{enumitem}
\usepackage{amsmath}
\usepackage{graphicx}
\graphicspath{{./}{../}}

\usepackage{url}
\usepackage{multirow}
\usepackage[table,dvipsnames]{xcolor}
\usepackage[most]{tcolorbox}
\usepackage{amssymb}
\usepackage{algpseudocode}
\usepackage{algorithm}
\usepackage{tabularx}
\usepackage{subcaption}
\usepackage{ragged2e}
\newcolumntype{L}[1]{>{\RaggedRight\arraybackslash\hsize=#1\hsize}X}

\definecolor{oursrow}{RGB}{229,229,247}

\definecolor{caseblue}{HTML}{EAF2FF}
\definecolor{casegreen}{HTML}{EAF7EF}
\definecolor{caseorange}{HTML}{FFF4E6}
\definecolor{casered}{HTML}{FDECEC}

\newtcolorbox{casebox}[2][]{
  enhanced,
  breakable,
  colback=#2,
  colframe=black!35,
  boxrule=0.5pt,
  arc=2pt,
  left=5pt,
  right=5pt,
  top=5pt,
  bottom=5pt,
  fonttitle=\bfseries,
  #1
}

\newcommand{\systemname}{\textsc{FinSkillOps}}
\newcommand{\evolvename}{\textsc{FinSkillOps}}

\title{\evolvename{}: A Self-Evolving Multi-Agent System for SEC Filing QA}

\author{
Yanzhang Ma\textsuperscript{1},
Zhenghan Tai\textsuperscript{1,3},
Hanwei Wu\textsuperscript{1,9},
Sizhe Guan\textsuperscript{1,9},
Jianliang Lei\textsuperscript{1},
Hailin He\textsuperscript{1},
\\ \bfseries
Chaolong Jiang\textsuperscript{1},
Jijun Chi\textsuperscript{3},
Tung Sum Thomas Kwok\textsuperscript{1,4},
Bohuai Xiao\textsuperscript{1},
Jingrui Tian\textsuperscript{2},
Xinlu Wu\textsuperscript{1},
\\ \bfseries
Xingao Zhan\textsuperscript{11},
Peng Lu\textsuperscript{7},
Muzhi Li\textsuperscript{5},
Yihong Wu\textsuperscript{7},
Liheng Ma\textsuperscript{1,2,8},
Sicheng Lyu\textsuperscript{1,2,8},
Tianshuo Yan\textsuperscript{12},
\\ \bfseries
Junhao Zhu\textsuperscript{10},
Yaqian Xu\textsuperscript{3},
Lei Ding\textsuperscript{1,6},
Yufei Cui\textsuperscript{2},
Ziquan Liu\textsuperscript{14},
Boyu Han\textsuperscript{1,13},
\\ \bfseries
Hengli Liu\textsuperscript{15},
Ling Zhou\textsuperscript{16},
Xinyu Wang\textsuperscript{1,2}%
\thanks{Contact: \texttt{xinyu.wang5@mail.mcgill.ca}}
\\[3pt]
{\small
\textsuperscript{1}SimpleWay.AI \quad
\textsuperscript{2}McGill University \quad
\textsuperscript{3}University of Toronto \quad
\textsuperscript{4}University of California, Los Angeles}
\\
{\small
\textsuperscript{5}The Chinese University of Hong Kong \quad
\textsuperscript{6}University of Manitoba \quad
\textsuperscript{7}Universit\'e de Montr\'eal \quad
\textsuperscript{8}Mila -- Quebec AI Institute}
\\
{\small
\textsuperscript{9}McMaster University \quad
\textsuperscript{10}Harvard University \quad
\textsuperscript{11}Monash University \quad
\textsuperscript{12}The University of Hong Kong}
\\
{\small
\textsuperscript{13}Stanford University \quad
\textsuperscript{14}Queen Mary University of London \quad
\textsuperscript{15}UE Capital Ltd. \quad
\textsuperscript{16}CG Matrix Technology Ltd.}
}

\begin{document}
\maketitle

\begin{abstract}
Financial QA systems are typically improved before deployment through better retrieval, prompting, or agent coordination, leaving their reliability behavior fixed thereafter. In practice, new SEC-filing questions repeatedly expose heterogeneous errors in period, entity, evidence use, and calculation. Existing self-improvement methods can turn failures into new behaviors, but offer limited control over where a correction should apply or which previously correct answers it may break. We therefore frame post-deployment improvement as controlled behavioral maintenance: recurring failures should become scoped skill patches, and each patch should earn deployment without introducing regressions. We instantiate this view in \systemname{}, a multi-agent system for SEC filing QA. \systemname{} derives reusable skills from evidence-grounded, typed failure diagnoses and governs them through targeted validation, protected-case regression checks, negative controls, and versioned replacement or retirement. Across six financial QA benchmarks, a single frozen skill registry achieves the highest verdict-weighted correctness and reference consistency among the evaluated systems. Evolved skills raise correctness from 3.70 to 4.55 on our enhanced benchmark. In a separate 12-round operational study, only six of 33 proposed skills are promoted, while the monitoring non-correct rate falls from 20.0\% to 12.5\%. These results establish controlled skill scope, admission, and lifecycle management as the foundation for reliable self-improvement.
\end{abstract}

\section{Introduction}

An analyst asking how a company's export-control exposure changed in fiscal
2025 needs an answer grounded in evidence for the correct fiscal year and
reporting entity.
Retrieval-augmented generation supports this by conditioning answers on
retrieved passages~\citep{lewis2020rag}.
Even with relevant evidence, a system can mix reporting periods, combine
figures from different entities, or compute a share inconsistent with the
cited amounts.
When these errors recur across filings and query styles, analysts must
repeatedly check the answers against their sources.

Financial QA systems combine components such as
domain-tuned models, structured indexing, reranking, multi-agent aggregation,
and debate~\citep{tian2024fingpt,cai-etal-2025-findebate,wang2024mixtureofagents,zhang2025pageindex},
along with domain prompts that specify period anchoring, explicit
denominators, and formulas for derived figures.
Self-improving agents update reusable behaviors from trajectories,
reflection, or execution
feedback~\citep{shinn2023reflexion,wang2023voyager,yang2026skillopt,song2026skillops}.
For SEC filing QA, period, entity, evidence, and arithmetic errors require
different corrections, even when the retrieved context is relevant.
We study how diagnosed failures can be converted into reusable instructions.
Each update needs a clear scope of application, an injection point in the
answering pipeline, and checks for regressions on a protected set.

\evolvename{} combines role-specialized SEC filing QA with an offline skill-update loop. Filings are curated into modality-tagged retrieval units; an orchestrator activates role-specialized analyst agents that decompose the question, retrieve evidence along complementary paths, and produce grounded drafts for synthesis.
The system maintains a registry of \emph{skills} for managing reusable natural-language instructions with explicit applicability conditions. Offline, failed answers are diagnosed against filing evidence. The system then drafts or revises a skill and evaluates it through targeted validation, protected-suite regression checks, and negative-control tests. Accepted updates are recorded in the registry with their originating failures and validation results. At serving time, applicable skills from a frozen snapshot guide query decomposition and final synthesis. Our contributions are:
\begin{itemize}[leftmargin=*,nosep]
\item We formulate post-deployment reliability in SEC filing QA as controlled skill maintenance: recurring failures become scoped behavioral updates whose benefits, regressions, and false activations are assessed before deployment.
\item We introduce \systemname{}, which derives evidence-grounded skills and manages their promotion, revision, replacement, and retirement through frozen registry snapshots.
\item Experiments on six benchmarks show improved correctness and reference consistency. In a twelve-round study, lifecycle management reduces the non-correct rate from 20.0\% to 12.5\%, admitting six of 33 proposed skills.
\end{itemize}

\section{Related Work}

\paragraph{Financial question answering over filings.}
Financial QA benchmarks have long targeted numerical and tabular reasoning
over disclosures~\citep{chen2021finqa,zhu2021tatqa,chen2022convfinqa}, and
recent systems improve filing-scale answering through domain-specific
models~\citep{tian2024fingpt}, multi-agent debate~\citep{cai-etal-2025-findebate},
agent aggregation~\citep{wang2024mixtureofagents}, structured document
indexing~\citep{zhang2025pageindex}, and end-to-end financial RAG
pipelines~\citep{FinSage}.
These systems fix their reliability behavior at build time.
When deployment exposes a repeated failure mode, the behavior that would
prevent it has to be written into a prompt or a component by hand.

\paragraph{Self-evolving agents and skill management.}
Agents can update reusable behaviors without retraining, through textual
skill editing~\citep{yang2026skillopt,yang2026autoskill,ni2026trace2skill,liu2026skillforge},
executable skill generation~\citep{wang2023voyager,lin2026museautoskill,zhang2026coevoskills},
and reflective memory~\citep{shinn2023reflexion,he-etal-2025-enabling}; recent
surveys organize this space by how skills are acquired, stored, and
governed~\citep{xu2026agentskills}.
Their update signal is typically a trajectory outcome or a scalar reward.
SEC filing QA needs attribution at the level of the claim that went wrong,
because period, entity, evidence, and calculation errors coexist under the
same failed answer, and it needs an admission decision that accounts for the
answers a new behavior might break.
\systemname{} supplies both: typed diagnosis grounded in retrieved filing
text, and promotion gated on targeted, protected, and negative-control
evidence.

\section{\evolvename{}}

\begin{figure*}[t]
    \centering
    \includegraphics[width=\textwidth]{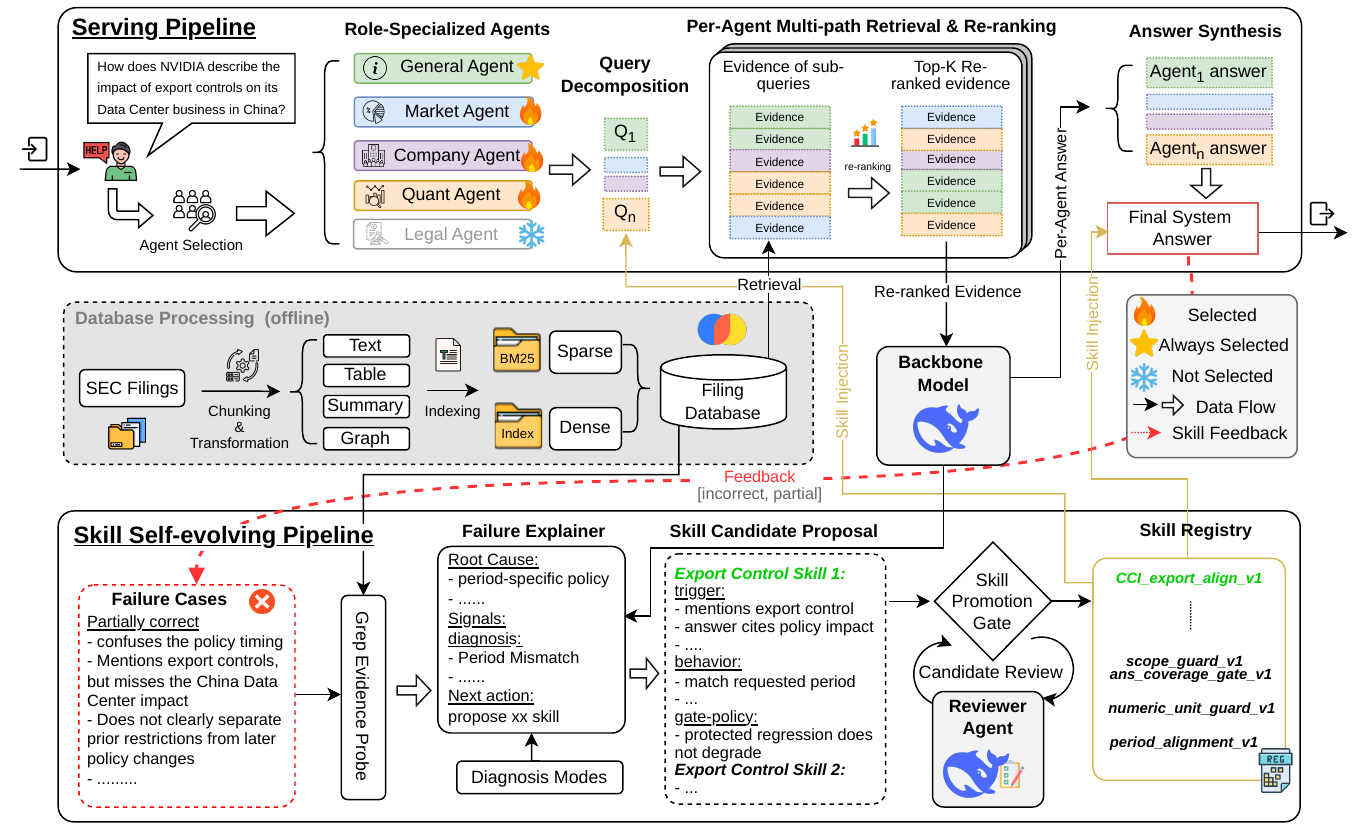}
    \caption{\evolvename{}: the serving pipeline (top) draws on curated filing
    indexes (middle), while the evolution loop (bottom) turns failed answers
    into gated registry updates that re-enter serving at the two marked
    injection points. The Reviewer Agent reviews candidate skills.}
    \label{fig:system}
\end{figure*}

\subsection{Task and System Overview}
\label{sec:formulation}

Let $\mathcal{D}$ be a corpus of SEC filings and $q$ a natural-language
question over $\mathcal{D}$.
The system retrieves an evidence set $E$ from $\mathcal{D}$ and produces an
answer $a$ conditioned on $(q,E)$; the reference answer is $a^{*}$.
A response succeeds when it covers the key points annotated for $a^{*}$
without contradicting them.
Diagnosis identifies root-cause modes in $\mathcal{M}$, such as
period mismatch, entity confusion, an unsupported claim, or a calculation
error. Judge error categories provide answer-level feedback; diagnosis uses
this feedback and inspected evidence to assign modes for clustering and
skill proposal. Appendix~\ref{app:revision-prompts} distinguishes the label
schemes and lists the eight diagnosis modes in the twelve-round record.

A \emph{skill} $s$ is a natural-language reliability instruction paired with
applicability conditions, false-trigger guards, a version, and a status.
$\mathcal{S}=\{s_1,\dots,s_n\}$ is the active skill set stored in the
registry, and $A_{\mathcal{S}}$ denotes the system using this set.
For a fixed serving configuration, it differs from the base system
$A_{\varnothing}$ only in the injected skill instructions.
The objective is to derive updates
$\mathcal{S}^{(t)}\!\rightarrow\!\mathcal{S}^{(t+1)}$ from failure feedback
that improve targeted validation performance and pass protected-suite and
applicability checks.

\paragraph{Serving pipeline.}
\label{sec:substrate}
Filings are parsed into narrative blocks, rendered tables, and section
summaries. Each retrieval unit retains its modality, section path, reporting
period, and filing entity as metadata.
An orchestrator activates a subset of five analyst roles (general,
quantitative, market, legal, and company) drawn from the principal
disclosure categories of SEC filings.
Each activated agent reads the section summaries and rewrites $q$ into
role-specific sub-queries using the filing's terminology. It then retrieves
evidence through sparse, dense, and summary paths. The orchestrator
synthesizes the agent drafts into $a$.
Selected skill instructions are injected at two points: query decomposition
to constrain evidence retrieval, and final synthesis to constrain answer
claims.
Answers that fail evaluation are recorded with their evidence and judge
feedback and enter the offline update path.
Serving, diagnosis, and skill proposal use DeepSeek-V3.2; answer evaluation on benchmarks uses Qwen3-Max.
Parsing, indexing, role prompts, and retrieval parameters are given in
Appendices~\ref{app:preprocess} and~\ref{app:substrate}.

\subsection{Failure-Driven Skill Evolution}
\label{sec:evolution}

Each round evaluates the current system on a batch $\mathcal{B}_t$ from
the evolution pool and collects failed instances as $\mathcal{F}_t$.
Recurring failures in this subset guide candidate skill updates
(Algorithm~\ref{alg:evolution}).

\paragraph{Failure diagnosis.}
\label{sec:diagnosis}
Diagnosis uses two LLM stages. The first reads the question, failed answer,
judge feedback, and retrieved context. It proposes explanations for the
failure and a search plan for checking them against the filing corpus.
Deterministic grep executes this plan. The second stage reads the returned
snippets and produces the diagnosis, supporting evidence, and a
propose / park / do-not-fix recommendation.
Prompt inputs and outputs for both stages are in
Appendix~\ref{app:revision-prompts}.
Inspecting the filing text helps distinguish errors that need different
repairs, such as choosing the wrong period and miscalculating with the
correct operands. Failures are grouped by mode; modes whose frequency
reaches a threshold $\tau$ proceed to skill proposal.

\paragraph{Skill proposal.}
\label{sec:proposal}
For each mode that clears the threshold, an LLM drafts a skill from the
representative failures and their supporting evidence.
A candidate specifies its applicability conditions, the behavior to apply,
excluded cases, and the serving step where its instruction is injected.
The proposer may revise a skill instead of adding one, keeping
the registry small and avoids conflicts.
Section~\ref{sec:case} illustrates a skill that aligns periods and recomputes
category shares from table evidence.

\paragraph{Runtime application.}
Diagnosis and proposal are offline and use reference-derived feedback.
At each injection point, the router selects active skills from the frozen
registry by checking their applicability conditions and false-trigger guards
against the query, corpus metadata, and available evidence features.
Before query decomposition, the evidence view consists of section
summaries. At final synthesis, retrieved passages are also available.
The selected behavior text is inserted into the corresponding prompt.
Reference answers, key-point labels, and judge feedback are not serving
inputs.

\subsection{Gated Promotion and Skill Lifecycle}
\label{sec:gate}

Each candidate is checked against the current active skill set
$\widehat{\mathcal{S}}$, including any updates accepted earlier in the round.
Let $\mathcal{S}'$ be the set after inserting or revising candidate $s'$.
\emph{Targeted validation} compares the two sets on $\mathcal{V}_m$, a
subset of instances with mode $m$ drawn from the evolution pool.
The \emph{protected suite} $\mathcal{P}$ contains previously correct
instances and high-risk cases covering period alignment, numerical
reasoning, entity resolution, and evidence conflicts.
Every Correct verdict on $\mathcal{P}$ under $\widehat{\mathcal{S}}$ must remain Correct
under $\mathcal{S}'$.
\emph{Negative controls} test for unwanted activation on no-op,
multi-period, and cross-company queries outside the skill's scope.

Promotion requires
 
\begin{equation}
\begin{aligned}
  &\mathrm{Acc}_{\mathcal{V}_m}\!(\mathcal{S}')
    > \mathrm{Acc}_{\mathcal{V}_m}\!(\widehat{\mathcal{S}}) \\
  &\qquad \wedge\;
  \mathrm{Reg}_{\mathcal{P}}\!(\mathcal{S}',\widehat{\mathcal{S}})
    = 0 \\
  &\qquad \wedge\; \mathrm{FalseTrig}(s')=0 ,
\end{aligned}
\label{eq:gate}
\end{equation}
where $\mathrm{Acc}$ is the fraction of Correct verdicts,
$\mathrm{Reg}_{\mathcal{P}}$ counts protected-suite verdicts that change
from Correct under $\widehat{\mathcal{S}}$ to non-Correct under $\mathcal{S}'$,
and $\mathrm{FalseTrig}$ counts activations on the declared negative
controls.
Candidates may also be refused or deferred for insufficient benefit,
unstable triggering, or conflict with an active skill
(Appendix~\ref{app:revision-prompts}).

A promoted skill enters the registry with a version number.
A revised version supersedes the earlier one, which becomes inactive.
A skill whose behavior is absorbed by another and that stops activating
is retired after a removal test.
The registry retains each version's originating failures, validation results,
and protected-suite outcome, including those of inactive versions.

\section{Experimental Setup}

\subsection{Datasets}
We evaluate on three public and three in-house financial QA benchmarks. The public benchmarks are \texttt{FinanceBench}~\citep{islam2023financebench}, \texttt{SECQUE}~\citep{benyoash2025secque}, and \texttt{FinDER}~\citep{choi2025finder}. The in-house \texttt{Lotus} and \texttt{Zeekr} sets contain multi-entity QA pairs with annotated relevance labels from two automotive brands. We additionally construct an enhanced in-house benchmark of 114 questions (84 Zeekr, 20 Lotus, 10 NVIDIA) written to require finer-grained period alignment, entity and scope disambiguation, multi-hop evidence aggregation, and more precise numerical or regulatory reasoning. Every question is grounded in company SEC filings and paired with a reference answer and key points.

\subsection{Baselines and Metrics}
\label{sec:baselines}
We compare with financial RAG (FinSage~\citep{FinSage}), financial search (our FinGPT adaptation~\citep{tian2024fingpt}), multi-agent collaboration (MoA~\citep{wang2024mixtureofagents}, FinDebate~\citep{cai-etal-2025-findebate}), and a direct retrieve-and-generate baseline (Naive RAG). Details are in Appendix~\ref{app:baselines}.

A zero-shot Qwen3-Max judge scores each answer against its reference answer and key points. We report verdict-weighted correctness (\textbf{W-Corr.}; $0$--$5$), reference consistency (\textbf{Ref. Cons.}), and four complementary quality dimensions. On 150 sampled answers, the judge's agreement with two annotators from the author group was $\kappa=.820/.675$ for four-way verdicts and $\kappa=.827/.840$ for Correct versus non-correct labels. Full metric definitions, judge prompts, human-agreement results, and
retrieval settings are in Appendices~\ref{app:eval-protocol},
\ref{app:revision-human}, and~\ref{app:revision-config}.

\section{Results}
\label{sec:results}

\begin{table*}[th]
\centering
\scriptsize                                   
\setlength{\tabcolsep}{3pt}
\renewcommand{\arraystretch}{1.0}
\caption{Per-benchmark results. Best in \textbf{bold}, second best
\underline{underlined}.}
\label{tab:per_benchmark_results}

\newcommand{\benchhead}{\textbf{Method} & \textbf{W-Corr.} & \textbf{Info.} & \textbf{Reas.} & \textbf{Ref.\,Cons.} & \textbf{Clar.} & \textbf{Depth} \\}

\begin{minipage}[t]{0.49\textwidth}\centering
\textbf{Zeekr (in-house)}\\[2pt]
\begin{tabular}{l rrrrrr}
\toprule
\benchhead
\midrule
Naive RAG  & 2.67 & 2.47 & 2.96 & 3.56 & 4.37 & 2.50 \\
FinDebate  & 3.11 & 3.40 & 3.95 & 3.83 & \underline{4.65} & \underline{3.92} \\
MoA        & 3.40 & 3.57 & \underline{4.02} & 3.95 & 4.56 & \underline{3.92} \\
FinGPT     & 2.66 & 3.27 & 3.69 & 3.26 & 4.61 & 3.60 \\
FinSage    & 2.83 & 2.76 & 3.32 & 3.96 & 4.47 & 2.84 \\
Multi-agent init. & \underline{3.46} & \underline{3.72} & 3.97 & \underline{4.19} & 4.59 & 3.84 \\
\rowcolor{oursrow}
\evolvename{} full & \textbf{4.52} & \textbf{4.56} & \textbf{4.32} & \textbf{4.74} & \textbf{4.85} & \textbf{4.26} \\
\bottomrule
\end{tabular}
\end{minipage}
\hfill
\begin{minipage}[t]{0.49\textwidth}\centering
\textbf{FinanceBench (open)}\\[2pt]
\begin{tabular}{l rrrrrr}
\toprule
\benchhead
\midrule
Naive RAG  & 2.21 & 2.36 & 2.52 & 2.76 & 4.54 & 2.40 \\
FinDebate  & 3.17 & 3.57 & \underline{4.01} & 3.56 & \underline{4.68} & \underline{3.92} \\
MoA        & 3.05 & 3.43 & 3.79 & 3.36 & 4.55 & 3.64 \\
FinGPT     & \underline{3.35} & \textbf{4.10} & \textbf{4.35} & \underline{3.59} & \textbf{4.90} & \textbf{4.21} \\
FinSage    & 2.97 & 3.27 & 3.34 & 3.26 & 4.60 & 3.23 \\
Multi-agent init. & 3.15 & \underline{3.70} & 3.77 & 3.52 & 4.61 & 3.71 \\
\rowcolor{oursrow}
\evolvename{} full & \textbf{3.49} & 3.39 & 3.49 & \textbf{3.78} & 4.32 & 3.35 \\
\bottomrule
\end{tabular}
\end{minipage}

\vspace{0.18cm}

\begin{minipage}[t]{0.49\textwidth}\centering
\textbf{Lotus (in-house)}\\[2pt]
\begin{tabular}{l rrrrrr}
\toprule
\benchhead
\midrule
Naive RAG  & 2.93 & 2.98 & 3.45 & 3.68 & 4.50 & 3.05 \\
FinDebate  & 3.19 & 3.58 & 3.90 & 3.89 & \textbf{4.69} & \underline{3.96} \\
MoA        & \underline{3.67} & \textbf{4.03} & \textbf{4.14} & 4.11 & \underline{4.62} & \textbf{4.21} \\
FinGPT     & 2.30 & 2.82 & 3.13 & 3.49 & 4.43 & 3.07 \\
FinSage    & 3.08 & 3.26 & 3.75 & 3.99 & 4.53 & 3.48 \\
Multi-agent init. & 3.58 & 3.79 & \underline{3.91} & \underline{4.16} & 4.58 & 3.83 \\
\rowcolor{oursrow}
\evolvename{} full & \textbf{3.74} & \underline{3.91} & 3.83 & \textbf{4.20} & 4.50 & 3.80 \\
\bottomrule
\end{tabular}
\end{minipage}
\hfill
\begin{minipage}[t]{0.49\textwidth}\centering
\textbf{SECQUE (open)}\\[2pt]
\begin{tabular}{l rrrrrr}
\toprule
\benchhead
\midrule
Naive RAG  & 1.67 & 1.68 & 1.90 & 2.20 & 4.24 & 1.70 \\
FinDebate  & 2.34 & 2.88 & \textbf{3.26} & 2.73 & \textbf{4.36} & \textbf{3.17} \\
MoA        & 2.20 & 2.81 & 3.07 & 2.66 & 4.28 & 3.05 \\
FinGPT     & 1.92 & 2.49 & 2.81 & \underline{3.02} & 4.06 & 2.70 \\
FinSage    & 1.87 & 2.25 & 2.27 & 2.51 & 4.08 & 2.13 \\
Multi-agent init. & \underline{2.44} & \underline{2.92} & 2.98 & 2.95 & \underline{4.30} & 2.98 \\
\rowcolor{oursrow}
\evolvename{} full & \textbf{2.59} & \textbf{3.15} & \underline{3.17} & \textbf{3.14} & 4.17 & \underline{3.06} \\
\bottomrule
\end{tabular}
\end{minipage}

\vspace{0.18cm}

\begin{minipage}[t]{0.49\textwidth}\centering
\textbf{Enhanced in-house}\\[2pt]
\begin{tabular}{l rrrrrr}
\toprule
\benchhead
\midrule
Naive RAG  & 1.88 & 2.33 & 2.75 & 2.58 & 4.25 & 2.70 \\
FinDebate  & 2.23 & 2.65 & 2.58 & 3.30 & 3.55 & 2.35 \\
MoA        & 2.13 & 2.70 & 2.65 & 3.28 & 3.68 & 2.45 \\
FinGPT     & 2.43 & 2.43 & 2.60 & 3.40 & 3.70 & 2.60 \\
FinSage    & 3.55 & 3.50 & 3.00 & 3.60 & 4.55 & \underline{4.05} \\
Multi-agent init. & \underline{3.70} & \underline{3.85} & \underline{3.72} & \underline{3.95} & \underline{4.60} & 3.70 \\
\rowcolor{oursrow}
\evolvename{} full & \textbf{4.55} & \textbf{4.75} & \textbf{4.55} & \textbf{4.90} & \textbf{4.88} & \textbf{4.55} \\
\bottomrule
\end{tabular}
\end{minipage}
\hfill
\begin{minipage}[t]{0.49\textwidth}\centering
\textbf{FinDER (open)}\\[2pt]
\begin{tabular}{l rrrrrr}
\toprule
\benchhead
\midrule
Naive RAG  & 1.55 & 1.52 & 1.65 & 2.35 & 4.14 & 1.48 \\
FinDebate  & 2.61 & 2.68 & \textbf{3.32} & 3.25 & \textbf{4.44} & \textbf{3.24} \\
MoA        & 2.66 & 2.70 & 3.24 & \underline{3.58} & 4.25 & \underline{3.10} \\
FinGPT     & 2.66 & 2.56 & 3.04 & 3.32 & \underline{4.35} & 2.96 \\
FinSage    & 2.23 & 1.93 & 2.13 & 3.08 & 3.99 & 1.89 \\
Multi-agent init. & \underline{2.68} & \underline{2.83} & 3.15 & 3.56 & 4.25 & 3.01 \\
\rowcolor{oursrow}
\evolvename{} full & \textbf{2.88} & \textbf{2.85} & \underline{3.30}  & \textbf{3.83} & 3.99 & 2.83 \\
\bottomrule
\end{tabular}
\end{minipage}

\end{table*}

\subsection{Overall Answer Quality}
\label{sec:main_results}

Using the same frozen registry across datasets, \evolvename{} achieves the
highest W-Corr. and Ref. Cons. on all six benchmarks
(Table~\ref{tab:per_benchmark_results}). The advantage is larger on Zeekr and
the enhanced in-house benchmark than on the public benchmarks and Lotus.
Results on the other quality dimensions are mixed: FinGPT leads on four
dimensions on FinanceBench, and \evolvename{} ties for the lowest clarity on
FinDER.
\evolvename{} also scores above the adapted AutoSkill and SkillOpt methods
on average across four benchmarks (Table~\ref{tab:skill-adapter-summary}).

\subsection{Component Contributions}
\label{sec:ablation}

\begin{table}[t]
\centering
\scriptsize
\setlength{\tabcolsep}{2pt}
\captionsetup{skip=4pt}
\caption{\textbf{Component ablations}. Parentheses give Full-minus-ablation
W-Corr. differences. Target Val.\ $\Delta$Acc and Reg. are in-house measures;
FB is FinanceBench. Definitions and paired CIs: Appendix~\ref{app:revision-config}.}
\label{tab:ablation}
\begin{tabular*}{\columnwidth}{@{}l@{\extracolsep{\fill}}rrrrr@{}}
\toprule
\multirow{2}{*}{\textbf{Method}} & \multirow{2}{*}{\shortstack{\textbf{Target Val.}\\$\boldsymbol{\Delta}$\textbf{Acc}}} & \multicolumn{3}{c}{\textbf{W-Corr. ($\Delta$)}} & \multirow{2}{*}{\textbf{Reg.}} \\
\cmidrule(lr){3-5}
 & & In-house & FB & FinDER & \\
\midrule
\textbf{Full \evolvename{}} & +0.62 & \textbf{4.55} & \textbf{3.49} & \textbf{2.88} & 0 \\
\addlinespace[2pt]
w/o Skill Evolution & +0.00 & 3.70 & 3.15 & 2.68 & 0 \\[-1pt]
 & & $(+0.85)$ & $(+0.34)$ & $(+0.20)$ & \\
\addlinespace[2pt]
w/o Protected Gate & +0.78 & 4.32 & 3.38 & 2.79 & 3 \\[-1pt]
 & & $(+0.23)$ & $(+0.11)$ & $(+0.09)$ & \\
\addlinespace[2pt]
w/o Failure Taxonomy & +0.31 & 4.08 & 3.28 & 2.70 & 0 \\[-1pt]
 & & $(+0.47)$ & $(+0.21)$ & $(+0.18)$ & \\
\bottomrule
\end{tabular*}
\end{table}

\begin{table}[t]
\centering
\footnotesize
\setlength{\tabcolsep}{2pt}
\captionsetup{skip=4pt}
\caption{\textbf{Skill activation and removal}. W-Corr. and $\Delta$ are
in-house; $\Delta=4.55-\text{W-Corr.}$. Details and CIs:
Appendix~\ref{app:revision-config}.}
\label{tab:single-skill-ablation}
\begin{tabular*}{\columnwidth}{@{}l@{\extracolsep{\fill}}rrrrr@{}}
\toprule
\multirow{2}{*}{\textbf{Skill}} & \multicolumn{3}{c}{\textbf{Activations}} & \multicolumn{2}{c}{\textbf{Removal}} \\
\cmidrule(lr){2-4}\cmidrule(l){5-6}
 & In-h. & FB & FinDER & W-Corr. & $\Delta$ \\
\midrule
Period alignment v2 & 31 & 22 & 17 & 4.21 & +0.34 \\
Numeric unit guard v1 & 24 & 19 & 15 & 4.34 & +0.21 \\
Answer coverage gate v1 & 27 & 14 & 12 & 4.37 & +0.18 \\
Scope guard v1 & 19 & 6 & 5 & 4.43 & +0.12 \\
\bottomrule
\end{tabular*}
\end{table}

\begin{table}[t]
\centering
\footnotesize
\setlength{\tabcolsep}{2.5pt}
\captionsetup{skip=4pt}
\caption{\textbf{Comparison with adapted skill-evolution methods.}
Means equally weight Zeekr, Lotus, FinanceBench, and FinDER.
$\Delta$ is method minus init.; Wins counts benchmarks above init.
Adapter protocols and full results: Appendix~\ref{app:skill-adapters}.}
\label{tab:skill-adapter-summary}
\begin{tabular*}{\columnwidth}{@{}l@{\extracolsep{\fill}}rrr@{}}
\toprule
Method & Avg. W-Corr. & $\Delta$ vs. init & Wins \\
\midrule
AutoSkill & 2.00 & $-1.22$ & 0/4 \\
SkillOpt & 2.66 & $-0.56$ & 0/4 \\
Multi-agent init. & 3.22 & -- & 0/4 \\
\textbf{\evolvename{}} & \textbf{3.66} & $\boldsymbol{+0.44}$ & \textbf{4/4} \\
\bottomrule
\end{tabular*}
\end{table}


Table~\ref{tab:ablation} reports component ablations on three benchmarks.
Removing Skill Evolution causes the largest W-Corr. drop on all three.
In the in-house experiment, the ungated condition records larger gains
on its candidates' targeted sets but admits protected-case regressions.
On FinanceBench and FinDER, the paired intervals for Skill Evolution and
Failure Taxonomy exclude zero; the gate intervals include zero.
Among the individual skills in Table~\ref{tab:single-skill-ablation}, removing
period alignment causes the largest score drop. The Scope Guard interval
includes zero. Appendix~\ref{app:paired-ci} reports evaluation details.

\subsection{Evolution Dynamics and Lifecycle}
\label{sec:evolution_analysis}

\begin{figure}[tb]
  \centering
  \includegraphics[width=\columnwidth]{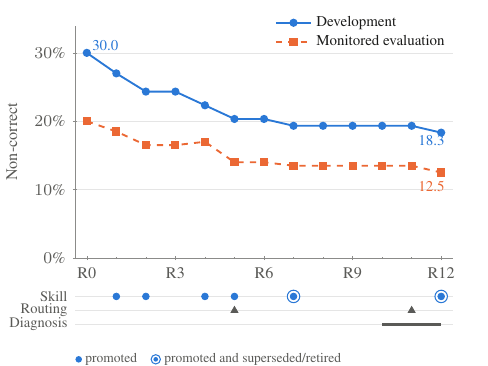}
  \caption{Non-correct rates over R0--R12. The lanes mark skill promotions
  and retirements, routing changes, and the diagnosis upgrade.}
  \label{fig:evolution}
\end{figure}

To study skill evolution over time, we ran twelve update rounds with disjoint evolution (300 questions), protected (120), and monitoring (200) sets. The evolution set supplied failures and targeted validation cases, the protected set gated promotions, and the monitoring set was evaluated after every round. As shown in Figure~\ref{fig:evolution}, the non-correct rate fell from 30.0\% to 18.3\% on the evolution set and from 20.0\% to 12.5\% on the monitoring set.

Only six of 33 proposed skills were promoted. Five candidates failed protected-set checks, exposing seven regressions, while post-round evaluation found no protected regressions in the deployed registry. Two rejected candidates illustrate the role of the gate: Table Row Lock prevented valid cross-row aggregation, while Multi-Hop Decomposition improved targeted cases but over-decomposed simple questions and diverted retrieval. A fusion skill was likewise withheld until its trigger became stable after a diagnosis update.

\begin{table*}[t]
\centering
\small
\caption{Scenarios and answer repairs. Appendix~\ref{app:case_study}
(Cases A and C) gives the case context and skill behavior.}
\label{tab:answer-repairs}

\definecolor{repairink}{HTML}{314A60}
\definecolor{repairband}{HTML}{EFF3F6}
\definecolor{headA}{HTML}{F7E0E0}   
\definecolor{headB}{HTML}{DEE9F5}   
\definecolor{headC}{HTML}{DFEFE0}   

\newcolumntype{Y}[1]{%
  >{\hsize=#1\hsize\linewidth=\hsize\raggedright\arraybackslash}X}

\setlength{\tabcolsep}{5pt}
\renewcommand{\arraystretch}{1.15}
\newcommand{\hdstrut}{\rule[-1.0ex]{0pt}{3.4ex}}
\arrayrulecolor{repairink}
\begin{tabularx}{\textwidth}{@{}Y{0.76} c Y{1.02} c Y{1.22}@{}}
\toprule
\cellcolor{headA}\hdstrut\textcolor{repairink}{\textbf{Earlier answer}} &
$\rightarrow$ &
\cellcolor{headB}\hdstrut\textcolor{repairink}{\textbf{Skill-guided repair}} &
$\rightarrow$ &
\cellcolor{headC}\hdstrut\textcolor{repairink}{\textbf{Revised answer}} \\
\midrule
\rowcolor{repairband}
\multicolumn{5}{@{}l@{}}{\strut\textcolor{repairink}{\textbf{Scenario A}\quad
NVIDIA export controls and China Data Center business (FY2025)}} \\
\addlinespace[5pt]
Later H20 restrictions and FY2026 charges were mixed into the FY2025 account.
\par\smallskip
\textit{Failure: period mismatch.}
& &
\textbf{Period Align}
\par\smallskip
Select evidence matching the requested fiscal year when retrieved policy disclosures conflict.
& &
NVIDIA offered compliant China-market products. China Data Center revenue
grew in FY2025, while its share of total Data Center revenue remained below
the pre-control level.
\par\smallskip
\textbf{\textsc{Partial} $\rightarrow$ \textsc{Correct}}
\par
4/4 key points matched.
\\
\addlinespace[8pt]
\rowcolor{repairband}
\multicolumn{5}{@{}l@{}}{\strut\textcolor{repairink}{\textbf{Scenario C}\quad
Comparing category shares across 2021 and 2023}} \\
\addlinespace[5pt]
The 2023 shares were 60.7\%, 36.8\%, and 2.5\%, inconsistent with the cited category amounts and total.
\par\smallskip
\textit{Failure: inconsistent shares.}
& &
\textbf{Multi-period alignment repair}
\par\smallskip
Align each numerator and denominator to the same period, recompute the shares, and replace conflicting draft percentages.
& &
The revised shares are \textbf{64.3\%, 30.8\%, and 4.9\%}.
\par\smallskip
\textbf{4/4 key points matched.}
\\
\bottomrule
\end{tabularx}
\arrayrulecolor{black}
\end{table*}



The registry evolved through consolidation rather than accumulation. It retained four active skills from R5 to R12 while its static skill text shrank from 421 to 347 tokens. At R7, an expanded period/entity skill replaced its earlier version; at R12, the fusion skill absorbed the slot-coverage fallback, allowing the dormant skill to be retired. The run also included a routing-priority change at R5 and a diagnosis update before R12. Appendix~\ref{app:revision-evolution} provides the complete candidate and version history.

We repeated the evolution process for six rounds on FinanceBench and SECQUE. Monitoring-set W-Corr. increased by 0.40 and 0.25, respectively, and both runs ended with two active skills. Appendix~\ref{app:public-evolution} reports their histories and cross-registry evaluation.

\subsection{Examples of Answer Repair}
\label{sec:case}

Table~\ref{tab:answer-repairs} summarizes two recorded repairs.
In the numerical case, the static prompt already requires period anchoring,
explicit denominators, and formulas. The added skill specifies what to do
when a draft percentage conflicts with the cited amounts: align the
operands to the same period and recompute the share.

\subsection{Serving Cost}
\label{sec:efficiency}

\begin{figure}[!htbp]
  \centering
  \includegraphics[width=\columnwidth]{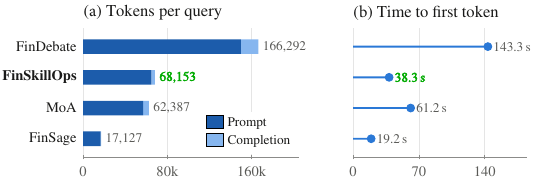}
  \caption{Serving cost by system: \textbf{(a)} prompt and completion tokens
  per query; \textbf{(b)} time to first token. Both panels share the row
  ordering.}
  \label{fig:serving-cost}
\end{figure}

Figure~\ref{fig:serving-cost} compares token use and time to first token.
\evolvename{} uses a token budget comparable to MoA and lower than
FinDebate's, with a shorter time to first token than both.
FinSage uses the fewest tokens and has the shortest time to first token.
\evolvename{} uses roughly four times as many tokens as FinSage.
Prompt tokens dominate usage across systems. Skill text is a small part of
this input, although its marginal serving cost was not measured separately.

\section{Conclusion}
\evolvename{} treats reliability in SEC filing QA as something a deployed
system maintains. Failures are attributed to typed error modes against
retrieved filing text, turned into natural-language skills with explicit
applicability conditions, and admitted only after targeted validation, a
protected-suite check, and negative controls. The system records the highest
correctness and reference consistency among the evaluated alternatives on six
benchmarks, and the twelve-round study shows the registry both growing and
shrinking.

\clearpage
\section{Limitations}
Our evidence covers SEC filing QA. Appendix~\ref{app:skill-adapters} reports
adaptations of other skill-evolution methods; comparisons under matched
evolution budgets and other financial workflows remain future work.
The twelve-round study is an operational trajectory that also changed routing
and diagnosis, and its evaluation set was monitored every round, so it
measures how the deployment improved rather than the isolated effect of skill
text, and it does not establish blind generalization.
The qualitative cases illustrate specific repairs and unresolved errors
through summarized answers and skill behavior. Available candidate
artifacts, admission logs, and split manifests are incomplete, limiting
verification of gate decisions and item-level separation between evolution
inputs and the benchmark or ablation questions.
Conservative gates slow adaptation: 27 of 33 candidates were refused,
deferred, or withdrawn, and admission requires evaluation runs on the
targeted, protected, and negative-control sets.
The financial-expert review route was not staffed, so admission decisions
rested on the automatic checks, and the judge-agreement study uses annotators
from the author group rather than independent domain experts.

\bibliography{custom}

\appendix
\section{SEC Filing Pre-processing}
\label{app:preprocess}
\systemname{} ingests raw SEC filings through a pre-processing pipeline, following the curation design of VeritasFi~\citep{veritasfi2025}.
The pipeline produces a retrieval-ready, modality-aware corpus in four stages.

\paragraph{Multi-modal parsing.}
Each filing is parsed with MinerU~\citep{wang2024mineru} into ordered, modality-tagged blocks (text, table, figure) annotated with their section hierarchy, i.e., the title path from the document root to each block.
Narrative blocks are kept directly.
Tables are handled on a dedicated pathway: each table region is converted by an LLM into a faithful textual/HTML rendering together with a short title summary, then stored in a separate table index so that tabular evidence is retrievable alongside narrative text.
Figures are summarized into descriptive captions.

\paragraph{Semantic enhancement.}
Curated chunks undergo three refinements that make them coherent and self-contained when retrieved in isolation.
(i)~\emph{Near-duplicate removal}: a chunk is discarded when its similarity to a retained chunk exceeds a threshold,
\begin{equation}
\mathcal{C}' = \mathcal{C}\setminus\{\,c_j\in\mathcal{C}\mid \exists\, i\neq j:\ \mathrm{sim}(c_i,c_j)>\tau_{\mathrm{sim}}\,\},
\end{equation}
implemented with a lexical TF--IDF / MinHash similarity at $\tau_{\mathrm{sim}}{=}0.95$.
(ii)~\emph{Coreference resolution}: pronouns are replaced with their explicit antecedents by an LLM, since chunks retrieved in isolation otherwise lack the context to resolve references.
(iii)~\emph{Metadata augmentation}: a section-level summary is generated and attached to every chunk as a shared contextual anchor for high-level queries.

\paragraph{Indexing.}
Enhanced chunks are split to a fixed length ($\sim$256 words) and indexed in parallel: dense embeddings via \texttt{bge-m3}~\citep{chen2024bgem3} stored in a Chroma vector store~\citep{chroma2025}, a BM25 sparse inverted index, and a separate section-summary embedding index that backs the summary retrieval path used by multi-path retrieval (Appendix~\ref{app:substrate}).
A lightweight document registry tracks per-company filings and their reporting periods.

\section{Multi-Agent Substrate Details}
\label{app:substrate}

This appendix documents the fixed SEC filing QA substrate used in our
experiments. The substrate provides the filing parser, retrieval indexes,
agent roles, query decomposition procedure, and answer-synthesis pipeline.
The model and retrieval backbone are fixed during evolution. The operational study additionally changes skill routing and diagnosis, as recorded in Appendix~\ref{app:revision-evolution}; these interventions are distinct from the fixed-configuration ablations.

\subsection{SEC Filing QA Substrate}

Given a question~$q$ and a pre-indexed filing corpus~$\mathcal{D}$, the
substrate retrieves evidence~$E$ and generates an answer~$a$ following a
retrieval-augmented generation pattern~\citep{lewis2020rag}.


\paragraph{Role-specialized analysis.}
An orchestrator receives the user question and activates a subset of five
analyst agents: general, quantitative, market, legal, and company. The
general agent is always active, while specialist agents are selected when
their expertise is relevant to the question. Each activated agent forms a
role-conditioned view of the question and uses it to guide retrieval and
draft generation. The five agent roles are summarized in
Table~\ref{tab:agent-roles}, and a representative quantitative-agent prompt
is shown below.

\paragraph{Database-aware query decomposition.}
Before generating sub-queries, each agent is exposed to a lightweight local
view of the filing database obtained from section-level summary retrieval.
The agent then rewrites the original question into role-specific,
data-seeking sub-queries grounded in the available filing structure and
terminology.

\paragraph{Multi-path retrieval.}
For each sub-query, the substrate retrieves candidate evidence from sparse,
dense, and summary-level retrieval paths. Retrieved candidates are ranked and
filtered to produce the evidence used by each agent. This multi-path design
improves coverage across exact financial terms, semantically related
disclosures, and section-level context.

\paragraph{Answer synthesis.}
Each activated agent generates a grounded draft from its retrieved evidence.
The orchestrator then combines the agent drafts and evidence into a final
synthesis prompt. Selected behavior text from~$\mathcal{S}$ is injected into query decomposition and final synthesis, allowing reliability behaviors such as period
alignment, numerical verification, and unsupported-answer blocking to affect
the final answer without changing the underlying substrate.

\subsection{Agent Role Definitions and Prompt Example}

Table~\ref{tab:agent-roles} summarizes the five analyst agents in the
substrate. Each agent is implemented via a role-specific prompt that defines
its responsibility, retrieval scope, and output style. The general agent
handles broad or conceptual questions and is activated for every query. The
remaining specialist agents are activated selectively by the orchestrator.

\begin{table*}[t]
\centering
\small
\caption{Summary of the five analyst agents in the substrate.}
\label{tab:agent-roles}
\begin{tabularx}{\textwidth}{@{}lXX@{}}
\toprule
\textbf{Agent} & \textbf{Focus} & \textbf{Typical triggers} \\
\midrule
General
& Overarching affairs, simple factual questions, and conceptual questions
& Always activated \\[4pt]
Quantitative
& Financial metrics, numerical reasoning, and financial statement analysis
& Revenue, profit, cash flow, valuation, financial ratios, or calculation-intensive queries \\[4pt]
Market
& Industry context, market share, competitive positioning, and business model
& Industry trends, competitor comparisons, customer segments, or market positioning \\[4pt]
Legal
& Legal proceedings, compliance, risk disclosures, governance, and transaction structure
& Regulation, governance, legal disputes, deal terms, or risk allocation \\[4pt]
Company
& Business structure, operations, ownership history, management, and milestones
& Company history, operating structure, fundraising, ownership, or management background \\
\bottomrule
\end{tabularx}
\end{table*}

Below we reproduce the prompt for the \emph{Quantitative Agent} as a
representative example. The prompt is split into a role description, a
query-decomposition phase, and an answer-generation phase. The static prompts already contain general grounding, period, and unit rules. Evolved skills add or revise triggered behaviors through the separate registry~$\mathcal{S}$.

\begin{tcolorbox}[
title={\textbf{Quantitative Agent -- Role Description}},
colback=gray!5,
colframe=gray!60,
fonttitle=\small\bfseries,
fontupper=\scriptsize,
breakable,
left=4pt, right=4pt, top=4pt, bottom=4pt
]
\textbf{Summary:} Focuses on financial statements and accounting-related
questions; produces concise answers with explicit units and periods.

\medskip
\textbf{Responsibilities:}
\begin{itemize}\setlength\itemsep{0pt}
\item Retrieve and interpret balance sheet, income statement, and cash flow
disclosures.
\item Extract financial figures with their units, periods, and source
context.
\item Identify when the retrieved evidence does not contain the requested
financial information.
\end{itemize}

\medskip
\textbf{Typical triggers:}
Questions about revenue, profit, cash flow, valuation, financial statement
items, accounting items, or calculation-intensive comparisons.

\medskip
\textbf{Exclusions:}
Pure market-background questions, legal/compliance assessments, or company
history unless they directly affect financial interpretation.
\end{tcolorbox}

\begin{tcolorbox}[
title={\textbf{Quantitative Agent -- Phase 1: Query Decomposition}},
colback=blue!3,
colframe=blue!40,
fonttitle=\small\bfseries,
fontupper=\scriptsize,
breakable,
left=4pt, right=4pt, top=4pt, bottom=4pt
]
You are a Quantitative Analysis Specialist. Your job is to rewrite the user's
question into atomic, searchable, data-seeking sub-questions related to
financial information in SEC filings.

\smallskip
Focus on the following areas when they are relevant to the user's question:
\begin{itemize}\setlength\itemsep{0pt}
\item Balance sheet items, including assets, liabilities, and equity.
\item Income statement items, including revenue, costs, expenses, and profit.
\item Cash flow statement items, including operating, investing, and
financing cash flows.
\item Financial ratios, accounting quantities, and numerical comparisons.
\item Accounting policies only when they are directly relevant to the
requested financial information.
\end{itemize}

\smallskip
Use the provided section summaries to understand what information may be
available in the filing database and to choose search terms that match the
filing structure.

\smallskip
Relevant section summaries: \texttt{{title summaries}}

\smallskip
Return only a JSON array of strings.\
User question: \texttt{{question}} \quad History: \texttt{{history}}
\end{tcolorbox}

\begin{tcolorbox}[
title={\textbf{Quantitative Agent -- Phase 2: Answer Generation}},
colback=green!3,
colframe=green!40,
fonttitle=\small\bfseries,
fontupper=\scriptsize,
breakable,
left=4pt, right=4pt, top=4pt, bottom=4pt
]
You are a Quantitative Analysis Specialist. Answer the user's question using
the provided evidence from SEC filings.

\smallskip
When relevant, include:
\begin{itemize}\setlength\itemsep{0pt}
\item The requested financial figures.
\item Units, periods, and company names.
\item Brief computations when the answer requires a calculation.
\item Source context from the provided evidence.
\item A brief note when the provided evidence does not contain enough
information to answer the question.
\end{itemize}

\smallskip
Question: \texttt{{question}} \quad
History: \texttt{{history}} \quad
Evidence: \texttt{{evidence}}

\smallskip
Return a concise answer grounded in the provided evidence.
\end{tcolorbox}

\section{Skill Evolution Algorithm}

Algorithm~\ref{alg:evolution} summarizes the candidate-update loop, including
frequency screening. The proposal ledger also counts suggestions rejected
or deferred before formal gating.
We use $\mathcal{E}$ to denote the evolution pool used for failure
diagnosis and targeted validation.
At evolution step~$t$, the system samples a batch $\mathcal{B}_t\subset
\mathcal{E}$ and runs the current skill-conditioned agent
$A_{\mathcal{S}^{(t)}}$.
Failed instances are collected as $\mathcal{F}_t$.
For each typed error mode~$m$, $\mathcal{V}_m\subset\mathcal{E}$ denotes a
targeted validation subset that shares mode~$m$.
The protected suite $\mathcal{P}$ contains previously passing and high-risk
instances for regression gating.
$\mathrm{Reg}_{\mathcal{P}}\!(\mathcal{S}',\mathcal{S})$ counts protected-suite
verdicts that change from Correct under $\mathcal{S}$ to non-Correct under
the candidate skill set $\mathcal{S}'$.
 
\begin{algorithm}[ht]
\caption{Skill Evolution Loop}
\label{alg:evolution}
\begin{algorithmic}[1]
\Require $A_{\varnothing}$, $\mathcal{S}^{(0)}$, $\mathcal{E}$, $\mathcal{P}$, $T$, threshold $\tau$
\Ensure $\mathcal{S}^{(T)}$
\For{$t = 0, 1, \dots, T\!-\!1$}
  \State Run $A_{\mathcal{S}^{(t)}}$ on batch $\mathcal{B}_t$
  \State $\mathcal{F}_t \!\leftarrow$ failed instances in $\mathcal{B}_t$
  \For{$i \in \mathcal{F}_t$} \Comment{diagnose}
    \State Diagnose answer, judge feedback, and evidence
    \State Record diagnosis modes and supporting evidence
  \EndFor
  \State $\mathcal{C} \!\leftarrow \{m \!\in\! \mathcal{M} : \mathrm{freq}(m) \!\geq\! \tau\}$
  \State $\widehat{\mathcal{S}} \!\leftarrow \mathcal{S}^{(t)}$
  \For{$m \in \mathcal{C}$} \Comment{propose \& gate}
    \State Draft candidate $s'_m$ from $m$ + examples
    \State $\mathcal{S}' \!\leftarrow$ insert or revise $s'_m$ in $\widehat{\mathcal{S}}$
    \If{$\mathrm{PassGate}(s'_m,\mathcal{S}',\widehat{\mathcal{S}})$}
      \State $\widehat{\mathcal{S}} \!\leftarrow \mathcal{S}'$; log promotion
    \Else
      \State Log rejection or deferral
    \EndIf
  \EndFor
  \State $\mathcal{S}^{(t+1)}\!\leftarrow\widehat{\mathcal{S}}$
\EndFor
\State \Return $\mathcal{S}^{(T)}$
\end{algorithmic}
\end{algorithm}
Here \textsc{PassGate} applies Equation~\ref{eq:gate} and the screening conditions in Appendix~\ref{app:revision-prompts} against the current active registry. The working set $\widehat{\mathcal{S}}$ preserves previously accepted updates within a round. Replacement marks the superseded version inactive; historical entries remain in the registry.

\section{Baseline Details}
\label{app:baselines}
We compare against four representative systems, drawn from financial RAG, domain-tuned financial LLMs, and multi-agent collaboration. Table~\ref{tab:per_benchmark_results} reports our implementations or adaptations evaluated through the shared harness, rather than numbers transcribed from the original papers. The descriptions below identify the source designs and evaluation adaptations; full shared settings are in Appendix~\ref{app:revision-config}.

\begin{itemize}
    \item \textbf{FinSage} \citep{FinSage}: An end-to-end RAG system for financial filing QA built from three components: a multi-modal pre-processing pipeline that unifies text, tables, and figures into metadata-enriched chunks; a multi-path retrieval module combining BM25, dense, metadata, and HyDE retrievers with neighbor-chunk bundling; and a domain-specialized re-ranker fine-tuned via Direct Preference Optimization (DPO).
    \item \textbf{FinGPT} \citep{tian2024fingpt}: A single-agent, tool-augmented financial search system built on the OpenAI Agent SDK with Model Context Protocol (MCP) integration. Each question is classified by lexical pattern matching and routed through one of three paths: an \emph{MCP-first} path for numerical queries that attempts structured MCP tool calls before falling back to web search; an \emph{iterative research} path for complex queries, where a planner LLM decomposes the question into typed sub-questions executed in parallel via MCP or web search with gap detection and multi-round refinement; and a \emph{single web search} path for simple or qualitative queries via the OpenAI Responses API. To ensure a fair comparison under our strictly document-grounded benchmark, we restrict FinGPT's tool access at evaluation by disabling the Yahoo Finance, TradingView, and filesystem MCP servers, leaving SEC-EDGAR as the only active structured data source.
    \item \textbf{MoA (Mixture-of-Agents)} \citep{wang2024mixtureofagents}: A general-purpose multi-agent framework with a layered architecture. Each layer contains multiple LLM agents, and every agent conditions on all outputs from the previous layer to refine its response; a final aggregator synthesizes these outputs into a single answer. The method operates purely through prompting, without fine-tuning.
    \item \textbf{FinDebate} \citep{cai-etal-2025-findebate}: A multi-agent framework for financial analysis pairing domain-specific RAG (FinLang embeddings over ChromaDB with context-sensitive chunking) with five specialized analyst agents (earnings, market, sentiment, valuation, risk) run in parallel. A safe debate protocol then refines the draft through Trust, Skeptic, and Leader agents in a single round, with the pre-debate stance fixed and every addition anchored to verifiable references.
\end{itemize}

\section{Evaluation Protocol Details}
\label{app:eval-protocol}
This appendix details our LLM-as-judge evaluation protocol. The reference-based rubric adapts the answer-level correctness labels, fine-grained quality dimensions, and error-type taxonomy of \citet{jiang2026finrate}. We employ a single LLM judge (Qwen3-Max), prompted in a zero-shot manner. The judge grounds its assessment solely on the provided gold answer and key points, and is constrained to penalize only direct contradictions with the reference, treating additional correct information as acceptable. The requested judge output consists of one verdict, five dimensional scores, and an optional error attribution; no inter-judge fusion is applied.

\subsection{Verdict Labels}
\label{app:verdict-labels}
Each response is assigned exactly one verdict:
\begin{itemize}[leftmargin=1.2em,itemsep=0.15em,topsep=0.2em]
    \item \textbf{Correct}: all key points are covered, verbatim or through reasonable paraphrase, with no contradictions against the reference.
    \item \textbf{Partial}: the response remains contradiction-free but omits essential key points.
    \item \textbf{Incorrect}: the response directly conflicts with the gold answer or key points. Omitting reference content does not count as incorrect if the response supplies alternative correct information.
    \item \textbf{Failure}: the response is a refusal, is irrelevant, or is empty.
\end{itemize}
A fifth operational label, \textbf{Error}, flags failures in the evaluation process, such as failed judge calls or unparseable outputs.

\paragraph{Correctness aggregation.}
The verdict-to-score mapping is Correct $=5$, Partial $=3$, Incorrect $=1$, and Failure $=0$. For each reported run, $N$ includes all evaluation records, including generation failures, evaluation errors, and fallback-processed records. Using the final correctness score $c_i$ assigned by evaluator post-processing to record $i$, the aggregate is
\begin{equation}
\text{W-Corr.}=\frac{1}{N}\sum_{i=1}^{N}c_i.
\end{equation}
W-Corr. is a verdict-weighted score on $0$--$5$, distinct from the proportion of answers judged Correct. The twelve-round non-correct count sums Partial, Incorrect, and Failure rather than using this weighted score.

\subsection{Fine-Grained Quality Dimensions}
\label{app:dimensions}

Beyond the categorical verdict, each response is rated on five dimensions using a $1$--$5$ scale:

\begin{itemize}
\item \textbf{Information Coverage}: whether the answer includes every query-critical fact and constraint needed to address the question, without diluting it with irrelevant content.
\item \textbf{Reasoning Chain}: whether the answer links evidence to intermediate conclusions to the final answer, rather than merely restating the evidence.
\item \textbf{Reference Consistency (Ref.\,Cons.)}: whether claims agree with the supplied reference answer and key points. Because the judge does not receive the full retrieved evidence, this score does not independently verify support for additional claims.
\item \textbf{Clarity of Expression}: whether the answer is organized so that the main conclusion is easy to locate, with minimal redundancy.
\item \textbf{Analytical Depth}: whether the answer prioritizes and synthesizes relevant evidence to reach a decisive, query-directed outcome, rather than summarizing indiscriminately.
\end{itemize}

The five dimensions are scored independently of the verdict and of one another. Each dimension uses scores after evaluator post-processing and the same denominator $N$, including all evaluation records. Failures, evaluation errors, and fallback-processed records remain in these denominators.

\subsection{Judge Prompt}
\label{app:judge-prompt}

The historical judge prompt---including the verdict definitions, key-point checking rules, the error-type taxonomy, and the strict output template---is reproduced verbatim in Fig~\ref{fig:prompt_judge}. The prompt specifies the requested model output; aggregation and fallback are handled separately by the evaluation wrapper.

\begin{figure*}[t]
\begin{tcolorbox}[
    enhanced,
    width=\textwidth,
    colback=blue!3,
    colframe=blue!32,
    title=\textsc{\evolvename{}} LLM-as-Judge Evaluation Prompt,
    fonttitle=\bfseries,
    coltitle=black,
    fontupper=\footnotesize,
]
\footnotesize
\textbf{System role:}

You are an expert evaluator for financial Q\&A tasks with retrieved
evidence. Use ONLY the provided gold answer / key points as reference.
The generated answer may contain additional correct information beyond
the gold answer; without access to the full retrieved evidence, you
cannot determine whether such additions are unsubstantiated. Therefore,
only penalize DIRECT CONTRADICTIONS with the gold answer / key points,
and do not penalize information merely absent from them.

\vspace{0.5em}
\textbf{Verdict labels (use exactly one):}
\begin{itemize}[leftmargin=1.2em,itemsep=0.15em,topsep=0.2em]
    \item \textbf{Correct}: all key points covered (verbatim or via
    reasonable paraphrase); no contradictions; relevant and complete.
    May include additional correct information.

    \item \textbf{Partial}: some key points covered but essential ones
    missing; no contradictions; relevant but incomplete.

    \item \textbf{Incorrect}: contains direct contradictions or clear
    factual errors conflicting with the reference. Not marked incorrect
    for merely omitting reference content if alternative correct
    information is provided.

    \item \textbf{Failure}: refuses to answer, is irrelevant, or is empty.

    \item \textbf{Error}: API call failed or other unexpected error
    (excluded from reported statistics).
\end{itemize}

\vspace{0.5em}
\textbf{Key-point evaluation:}
Check each key point against the generated answer and label it as
PRESENT, PARTIAL, MISSING, or INCORRECT, allowing numeric/date/percent
equivalence and reasonable paraphrase.

\vspace{0.5em}
\textbf{Error-type taxonomy} (choose NONE if Correct):
\begin{itemize}[leftmargin=1.2em,itemsep=0.15em,topsep=0.2em]
    \item \textbf{B. Generation-related}: B1 hallucination
    (B1-1 numeric/categorical, B1-2 entity attribute, B1-3 comparative
    stance, B1-4 trend/trajectory); B2 contradicts evidence;
    B3 excessive inference; B4 evidence-fusion failure.

    \item \textbf{C. Finance numeric/semantic}: C1 numerical precision;
    C2 units and scales; C3 time mismatch; C4 computation logic.

    \item \textbf{D. Query/context}: D1 query misunderstanding
    (D1-1 intent, D1-2 entity, D1-3 metric); D2 context-window abuse.
\end{itemize}
Output one primary group and subtype when the verdict is not Correct,
plus up to two secondary subtypes; prefer the most causal error.

\vspace{0.5em}
\textbf{Output format (strict):}
The judge emits, in fixed order, an \emph{Analysis}, per--key-point
labels with a summary count, five \emph{Dimensional Scores}
(Information Coverage, Reasoning Chain, Factual Consistency, Clarity of
Expression, Analytical Depth; each an integer $1$--$5$), the
\emph{Error Type} attribution, and a single \emph{Verdict} label.
\end{tcolorbox}
\caption{
Prompt template for the \textsc{\evolvename{}} LLM-as-judge evaluator. A
single judge (Qwen3-Max) scores each generated answer against the gold
answer and extracted key points, producing a verdict, five
fine-grained quality scores, and an error-type attribution. Evaluation
fields (question, gold answer, generated answer, key points) are
injected per instance. The historical Error-exclusion instruction does not
govern aggregation: all evaluation records enter the reported denominators.
}
\label{fig:prompt_judge}
\end{figure*}

\begin{figure*}[t]
\begin{tcolorbox}[
    enhanced,
    width=\textwidth,
    colback=blue!3,
    colframe=blue!32,
    title=\textsc{\evolvename{}} \textsc{Draft} Prompt Constructor,
    fonttitle=\bfseries,
    coltitle=black,
    fontupper=\footnotesize,
]
\footnotesize
\textbf{Specialist prompt:}

You are the activated specialty module inside a routed financial analysis
system. The upstream router selected this specialist; operate strictly
within your declared domain scope and do not invent facts not present in
the provided evidence or tool results.

\vspace{0.5em}
\textbf{Input:}
\begin{itemize}[leftmargin=1.2em,itemsep=0.15em,topsep=0.2em]
    \item Specialist name and domain scope.
    \item Available tools and tool results.
    \item Domain rules and output template (specialist-specific).
    \item Sub-question, retrieved evidence, and recent chat history.
\end{itemize}

\vspace{0.5em}
\textbf{Task rules:}
\begin{itemize}[leftmargin=1.2em,itemsep=0.15em,topsep=0.2em]
    \item Produce a domain-scoped structured answer by reading the
    sub-question and retrieved evidence through the lens of this
    specialist's declared rules.

    \item Keep every claim grounded in the provided evidence or tool
    results only; if a requested fact is absent, state so explicitly
    rather than inferring it.

    \item When tool results provide more recent or broader coverage than
    retrieved filing evidence, treat tools as authoritative and state
    the precedence explicitly rather than silently mixing sources.

    \item When tools supplement rather than supersede filing evidence,
    combine both to show a trend.
\end{itemize}

\vspace{0.5em}
\textbf{Output format:}

Specialist-specific structured sections defined in the static domain
template: \emph{Key Findings} (data-anchored bullets with period and
value), \emph{Supporting Evidence} (table references or excerpts),
\emph{Computations} (formula and steps for any derived figure),
\emph{Policy Notes} (only when explicitly relevant to the question), and
\emph{Missing Data} (only items that block the answer). Sub-question
drafts are generated in parallel and merged into a single draft before
the downstream \textsc{Synthesize} operator.

\vspace{0.5em}
\textbf{Example domain rules (quant analysis specialist):}

Focus on auditable financial interpretation within the declared scope
(balance sheet, income statement, cash flow, accounting policies):
\begin{itemize}[leftmargin=1.2em,itemsep=0.15em,topsep=0.2em]
    \item \textbf{Period anchor}: quarterly data $\neq$ cumulative;
    identify the exact quarter, fiscal year, or TTM period the user
    requests $\rightarrow$ anchor every cited value to that period.

    \item \textbf{Units and scale}: always include currency, unit, and
    denominator; distinguish quarterly from annual totals
    $\rightarrow$ no implicit unit conversion or omission.

    \item \textbf{Tool-vs-filing precedence}: tools provide more recent
    market or financial updates than retrieved filings $\rightarrow$
    overwrite older evidence and state the precedence explicitly.

    \item \textbf{Computation context}: directly reported value vs.\
    derived calculation vs.\ cross-period comparison $\rightarrow$
    show formula and steps for any derived figure.

    \item \textbf{Evidence sufficiency}: missing data, ambiguous metric
    naming, or period mismatch $\rightarrow$ enumerate only items
    that block the answer; do not report absent sections that are
    irrelevant.

    \item \textbf{Out-of-scope guardrail}: company history, competitor
    landscape, or legal and compliance matters $\rightarrow$ do not
    answer; those belong to other routed specialists.
\end{itemize}
\end{tcolorbox}
\caption{
Prompt template for the \textsc{\evolvename{}} \textsc{Draft} operator. Each
specialist carries its own static prompt encoding domain scope, grounding
rules, and output sections; the shared \texttt{draft\_answer} helper
injects the sub-question, per-sub-question evidence, tool results, and
chat history at runtime. The quant-analyst prompt is shown as an example.
}
\label{fig:prompt_draft_fin}
\end{figure*}

\section{Case Studies of Skill Evolution}
\label{app:case_study}

These qualitative cases illustrate how diagnosed failures motivate scoped
repairs and where errors remain. Cases A and B retain the question IDs,
protected-set labels, and reported outcomes from the original R1--R3
analysis, separate from the twelve-round rerun. Case C provides an
additional comparison between a static requirement and a specific repair
procedure. Answers and skill descriptions below summarize the case records.

\subsection{Export-Control Period Alignment}

\begin{casebox}[title={Case A: From period leakage to a correct export-control answer}]{casegreen}
\textbf{Dataset / QID.} NVIDIA protected set, \texttt{qa\_kp\_000015}

\smallskip
\textbf{Question.}
How did NVIDIA describe the impact of export controls on its China Data Center
business in fiscal year 2025?

\smallskip
\textbf{Reference answer.}
NVIDIA stated that it expanded its Data Center product portfolio with offerings
that did not require a license or prior notification before each shipment, and
introduced new China-market products that did not require an export license.
China Data Center revenue increased in fiscal year 2025, but its share of total
Data Center revenue remained far below the level before the October 2023 export
controls.

\smallskip
\textbf{Earlier answer [Partial].}
The earlier answer focused on later H20 restrictions, 2026 fiscal-year charges,
and subsequent policy uncertainty. It therefore conflated later regulatory
events with the fiscal-year-2025 disclosure and missed the key point that
NVIDIA reported compliant China-market alternatives and fiscal-year-2025 China
Data Center revenue growth.

\smallskip
\textbf{Diagnosis summary.}
\begin{itemize}
  \item \textbf{Failure type:} Period Mismatch + Evidence Conflict.
  \item \textbf{Signal:} retrieved evidence contained both fiscal-year-2025
  disclosure and later export-control developments.
  \item \textbf{Root cause:} the answer selected later policy events and
  overwrote the period requested by the question.
\end{itemize}

\smallskip
\textbf{Skill summary.}
\begin{quote}\small
\textbf{Period Align.}
Trigger when a question asks for a specific fiscal year or policy period and
retrieved evidence contains temporally conflicting disclosures.
Prefer evidence whose filing period matches the question; route to review when
later policy events would change the answer.
\end{quote}

\smallskip
\textbf{Reported gate outcome.}
The original case description reports that Period Align passed targeted
validation and protected checks and was promoted.

\smallskip
\textbf{Later answer [Correct].}
FinSkillOps answered under the fiscal-year-2025 disclosure: NVIDIA expanded its
Data Center portfolio, offered products that did not require a license or prior
notification for each shipment, introduced China-market products not requiring
an export license, and reported that China Data Center revenue increased while
remaining far below its pre-October-2023 export-control share of Data Center
revenue.

\smallskip
\textbf{Original judge trace (R1--R3).}
The case improved from \textsc{Partial} in R1 to \textsc{Correct} in R2/R3
(4/4 key points matched).
\end{casebox}

\subsection{Remaining Limitation: Residual Risk Coverage}

\begin{casebox}[title={Case B: A failure that remains partially unresolved}]{caseorange}
\textbf{Dataset / QID.} Lotus protected set, \texttt{lotus\_gen\_08}

\smallskip
\textbf{Question.}
Does Lotus Technology still operate through a VIE structure?

\smallskip
\textbf{Reference answer.}
Lotus Technology is a Cayman Islands holding company operating through
subsidiaries in China and Europe. Its restructuring eliminated the previous VIE
structure, so it is not a continuing VIE operating structure, although the
holding-company structure still carries China-related dividend and cash-transfer
risks.

\smallskip
\textbf{System answer [Partial].}
The answer correctly states that Lotus no longer operates through a VIE
structure and explains that the prior VIE contractual arrangements were
terminated during the restructuring. However, it does not fully preserve the
remaining holding-company and China-related dividend/cash-transfer risk
component.

\smallskip
\textbf{Diagnosis summary.}
\begin{itemize}
  \item \textbf{Failure type:} Coverage Gap + Scope Mismatch.
  \item \textbf{Signal:} the answer resolves the binary VIE question but drops
  residual risk language tied to the post-restructuring holding-company
  structure.
  \item \textbf{Root cause:} the skill library focuses on detecting whether a
  VIE remains active, but does not yet enforce residual-risk coverage after a
  restructuring.
\end{itemize}

\smallskip
\textbf{Skill summary.}
\begin{quote}\small
\textbf{Risk Carryover.}
Trigger when a question asks whether a legacy structure has been eliminated but
the evidence also discusses continuing risk factors after restructuring.
Require the answer to separate ``structure eliminated'' from ``risk remains.''
\end{quote}

\smallskip
\textbf{Reported gate outcome.}
The original case description reports that Risk Carryover was not promoted
because the pattern was too narrow and positive/negative controls were
insufficient.

\smallskip
\textbf{Original judge trace (R1--R3).}
The case remained \textsc{Partial} from R1 to R3, with 2/5 key points matched
and the residual China-related risk coverage still incomplete.
\end{casebox}

\paragraph{Takeaway.}
These cases illustrate period-specific evidence selection and residual-risk coverage. One reports a transition from \textsc{Partial} to \textsc{Correct}, while the other remains \textsc{Partial}. The twelve-round version history in Appendix~\ref{app:revision-evolution} separately documents skill replacement and retirement.

\subsection{From Static Checks to a Specific Recalculation Rule}
\label{app:case-delta}

\begin{casebox}[title={Case C: Recomputing category shares}]{casegreen}
\textbf{Static rule.}
The static quantitative prompt already requests period anchoring,
explicit units and denominators, and formulas for derived figures.
The additional skill specifies how to handle a conflict between a
draft percentage and the cited category amount and same-period total.

\smallskip
\textbf{Failure example.}
In a 2021/2023 comparison (\texttt{qa\_kp\_136}), the earlier answer
gave 2023 category shares of 60.7\%, 36.8\%, and 2.5\%, reported as
inconsistent with the cited table's category amounts and total.

\smallskip
\textbf{Trigger and behavior summary.}
The skill, \texttt{multi\_period\_alignment\_repair}, applies
to multi-period comparisons with table evidence assessed as sufficient.
It requires aligning each numerator and denominator to the same period,
recomputing the category shares, and replacing draft percentages that
conflict with those operands. The stated trigger uses query intent,
source type, and evidence sufficiency; question IDs identify the cases
for reporting and are not trigger conditions.

\smallskip
\textbf{Sample-level outcomes.}
The revised answer gives 64.3\%, 30.8\%, and 4.9\%, with four of four
reference key points matched. The same skill is reported to activate
on a separate 2021/2022 comparison (\texttt{qa\_kp\_110}), which retains
four of four matched key points. The two examples illustrate a repaired
answer and an unchanged comparison answer under the reported skill behavior.
\end{casebox}

\clearpage

\section{Model and Retrieval Configuration}
\label{app:revision-config}

Table~\ref{tab:revision-config} specifies the generation, evaluation,
and retrieval components. The reranker is served through vLLM as a
ranking model, not a generative LLM call. Generation and evaluation use
different model families; this separation does not eliminate judge bias.
The additional GPT-5.5 assessment in Appendix~\ref{app:revision-human}
is an agreement check, not the judge for the main results.

\paragraph{Registry snapshot and evaluation sets.}
Accepted skill updates are carried forward in one shared registry
across evolution rounds. For Table~\ref{tab:per_benchmark_results},
the final registry snapshot is frozen, with the same active skill
versions available across all six benchmarks; query-dependent
triggering selects from this fixed set. Multi-agent init. in
Table~\ref{tab:per_benchmark_results} denotes the same multi-agent substrate
in its initial state before skill evolution. Table~\ref{tab:ablation} reports
W-Corr. on the enhanced in-house benchmark and supplementary
FinanceBench and FinDER ablations. These evaluations are separate from the twelve-round rerun in
Appendix~\ref{app:revision-evolution}.

\paragraph{Controlled component settings.}
The in-house component comparisons hold the generation backbone,
retrieval substrate, answer-format requirements, and evaluation judge fixed.
The four configurations use the same judge model and version, prompt,
and decoding settings on the 114-question enhanced in-house benchmark.
The no-skill condition disables the skill-evolution layer. The no-gate
condition uses the same evolution pool and evaluation inputs as Full
\evolvename{}. These controls describe the in-house ablation experiment, separately from
the routing and diagnosis updates in the twelve-round study.

\paragraph{Targeted validation gain.}
In the in-house experiment in Table~\ref{tab:ablation}, Target Val.\ $\Delta$ measures the absolute
increase in the fraction of Correct verdicts (0--1 scale) on each
promoted candidate's targeted validation set $\mathcal{V}_m$
(Equation~\ref{eq:gate}). Each increase compares the registry immediately
before that candidate's admission with the registry after admitting it,
evaluated on the same candidate-specific set. We macro-average these
increases with equal weight per promotion event. All in-house configurations draw
validation items from the same evolution pool. The condition without Skill
Evolution proposes no candidates; we report $\Delta=0$ by convention.

\paragraph{Paired confidence intervals.}
\label{app:paired-ci}
For each in-house component ablation $a$ in Table~\ref{tab:ablation},
saved outputs are matched to Full by question ID.
Let $\mathcal{I}_a$ contain questions with valid final verdicts in both
conditions, and $n_a=|\mathcal{I}_a|$.
We compute $d_i=c_i^{\mathrm{Full}}-c_i^a$ using the verdict scores
Correct/Partial/Incorrect/Failure $=5/3/1/0$.
Failure remains a valid zero score; records lacking a valid verdict in
either condition are excluded from this paired analysis without imputation.
We estimate the mean of $d_i$ over $\mathcal{I}_a$ and its two-sided
95\% bias-corrected and accelerated (BCa) bootstrap interval using 10,000
resamples and a fixed random seed. Each resample draws $n_a$ question
indices with replacement and applies the same indices to both conditions.
Resampling operates on saved outputs and captures variation across questions.
The intervals for Full minus each ablation are $[0.57,1.16]$ without Skill
Evolution, $[0.05,0.41]$ without the Protected Gate, and $[0.24,0.70]$
without the Failure Taxonomy.
The in-house component means retain all evaluation records
(Appendix~\ref{app:eval-protocol}); the intervals use complete pairs.
When verdicts are missing, the paired-set mean difference can differ from
the difference between the table means.

For FinanceBench, the reported 95\% paired intervals for Full minus
each ablation are $[0.12,0.56]$ without Skill Evolution,
$[-0.02,0.24]$ without the Protected Gate, and $[0.04,0.38]$ without
the Failure Taxonomy. For FinDER, the corresponding intervals are
$[0.06,0.34]$, $[-0.04,0.22]$, and $[0.03,0.33]$.

\paragraph{Individual skills in the in-house experiment.}
Table~\ref{tab:single-skill-ablation} gives activation counts for the in-house
(114 questions), FinanceBench (150), and FinDER evaluations. The removal
scores and paired intervals compare Full with each skill removed from the
final in-house registry. The paired 95\% intervals are $[0.18,0.50]$ for
period alignment v2, $[0.06,0.36]$ for numeric unit guard v1,
$[0.03,0.33]$ for answer coverage gate v1, and $[-0.02,0.26]$ for
scope guard v1.
Table~\ref{tab:inhouse-skill-records} gives the corresponding skill
identifiers, target modes, admission rounds, and activation entries.
Admission $\Delta$ measures the targeted Correct-rate gain when a candidate
enters the registry, whereas the removal comparison measures its contribution
in the final registry. The in-house skill records and removal comparisons
are separate from the twelve-round record in
Appendix~\ref{app:revision-evolution}.

\begin{table*}[t]
\centering
\small
\setlength{\tabcolsep}{5pt}
\begin{tabular}{@{}llcrr@{}}
\toprule
\textbf{Skill (version)} & \textbf{Target mode} & \textbf{Round} & \textbf{Activation /114} & \textbf{Admission $\Delta$Acc} \\
\midrule
\texttt{period\_alignment\_v1} & Period mismatch & R3 & -- & +0.55 \\
\texttt{period\_alignment\_v2} & Period mismatch & R7 & 31 & +0.70 \\
\texttt{numeric\_unit\_guard\_v1} & Calculation error & R5 & 24 & +0.58 \\
\texttt{ans\_coverage\_gate\_v1} & Unsupported claim & R8 & 27 & +0.66 \\
\texttt{scope\_guard\_v1} & Entity confusion & R10 & 19 & +0.61 \\
\texttt{slot\_coverage\_v1} & Unsupported claim & R6 & -- & +0.62 \\
\bottomrule
\end{tabular}
\caption{Skill records for the enhanced in-house experiment.
Admission $\Delta$Acc is the Correct-rate gain on a candidate's targeted
validation set relative to the pre-admission registry. Activation values
are reported over the 114-question evaluation; a dash denotes an unreported
value. Skill names in Table~\ref{tab:single-skill-ablation} are shortened for display.}
\label{tab:inhouse-skill-records}
\end{table*}

\begin{table}[t]
\centering
\small
\begin{tabularx}{\columnwidth}{@{}lX@{}}
\toprule
Component / setting & Value \\
\midrule
Analysts (five roles) & DeepSeek-V3.2 \\
Orchestration & DeepSeek-V3.2 \\
Decomposition / synthesis & DeepSeek-V3.2 \\
Diagnosis / skill revision & DeepSeek-V3.2 \\
Table / figure rendering & DeepSeek-V3.2 \\
Coreference resolution & DeepSeek-V3.2 \\
Main judge & Qwen3-Max, zero-shot \\
Dense embedding & \texttt{BAAI/}\allowbreak\texttt{bge-m3} \\
Reranker & \texttt{BAAI/}\allowbreak\texttt{bge-reranker-}\allowbreak\texttt{v2-gemma} \\
Sparse retrieval & BM25 \\
QA / diagnosis / proposal $T$ & $0.0 / 0.1 / 0.2$ \\
Generation \texttt{top\_p} & Provider default \\
Synthesis token limit & 4096 \\
Other generation limits & Provider default \\
Judge $T$ / \texttt{top\_p} & $0.0 / 0.95$ \\
Judge token limit & 2048 \\
Dense / BM25 / table $k$ & $10 / 10 / 8$ \\
Rerank $K$ / chunk size & $8$ / ${\approx}256$ words \\
\bottomrule
\end{tabularx}
\caption{Model, decoding, and retrieval settings. Skill proposal and
revision share the diagnosis backbone. Unspecified provider defaults
are not replaced with assumed numerical values.}
\label{tab:revision-config}
\end{table}

\section{Evolution Prompt Interfaces}
\label{app:revision-prompts}

The following specifies prompt inputs and outputs; it is a structured
description rather than a verbatim template. The analyst and judge
templates are provided in Appendices~\ref{app:substrate}
and~\ref{app:judge-prompt}.

\paragraph{Offline diagnosis and runtime applicability.}
Offline diagnosis attributes answer failures using reference-derived
feedback and inspected evidence. Runtime applicability detection uses
the query, corpus metadata, and evidence-derived features to select
skills from the frozen registry; reference key-point labels and judge
feedback are not runtime inputs. In the twelve-round diagnosis record,
\texttt{Diag-B4} denotes an
observed evidence-fusion failure after answer generation. Runtime
applicability concerns whether an evidence-fusion behavior is relevant
to the current query and available evidence.
Selected behavior text is injected into query decomposition and final
synthesis. Section-summary retrieval precedes decomposition, whereas
full multi-path retrieval follows it; skill selection cannot use
evidence available only after the relevant injection.
The reference-based evaluation judge is separate from evidence-grounding
instructions within answer generation.

\paragraph{Two-stage diagnosis.}
Stage one receives the question, answer excerpt, judge verdict and
analysis, key-point counts, and a preview of retrieved passages. It
returns a retrieval plan containing hypotheses, queries with purposes,
and facts to verify. Queries use filing phrases, table labels, period
markers, segment names, or metric aliases. Deterministic grep executes
the plan against the filing corpus; each hit records its source path,
match score, matched terms, and snippet. Stage two receives these hits
with the original inputs and plan, and returns a root-cause type,
inspected evidence, failure explanation, overfitting risk, and a
propose / park / do-not-fix recommendation. The specified diagnosis inputs
exclude protected-suite records.

\paragraph{Error labels and record identifiers.}
Judge categories describe answer-level errors against the reference
(Appendix~\ref{app:judge-prompt}). Diagnosis uses that feedback together
with inspected evidence to assign root-cause modes for failure grouping
and skill proposal. Table~\ref{tab:diagnosis-modes} lists the eight modes
in the twelve-round record. Outside the reproduced judge prompt, we use
role prefixes to distinguish code references:
\texttt{Judge-C1} means numerical precision, while \texttt{Diag-C1} means
period mismatch. The record's \texttt{Conflict-C1}--\texttt{Conflict-C6} identify
six governance events, grouped into trigger overlap, instruction
contradiction, functional redundancy, and ordering/priority conflicts.
The prefixes distinguish the paper's references to these separate schemes.

\begin{table}[t]
\centering
\small
\begin{tabularx}{\columnwidth}{@{}lX@{}}
\toprule
Code & Diagnosis mode \\
\midrule
\texttt{Diag-B1} & Unsupported claim \\
\texttt{Diag-B2} & Unresolved evidence conflict \\
\texttt{Diag-B3} & Cross-filing entity confusion \\
\texttt{Diag-B4} & Multi-evidence fusion failure \\
\texttt{Diag-C1} & Period mismatch \\
\texttt{Diag-C2} & Table-unit or row-semantics misreading \\
\texttt{Diag-D1} & Missing answer slots \\
\texttt{Diag-D2} & Context-priority error \\
\bottomrule
\end{tabularx}
\caption{Diagnosis labels listed in the R0--R12 record. The displayed
\texttt{Diag-} prefix distinguishes them from the judge taxonomy.}
\label{tab:diagnosis-modes}
\end{table}

\paragraph{Proposal and registry.}
The proposer receives questions, answer excerpts, judge analyses, and
missing/incorrect key-point labels, but not the full reference answer
text. These labels supply reference-derived, answer-specific feedback.
The registry is a flat list of JSON entries with identifier, version,
status, trigger conditions, evidence signals, false-trigger guards,
natural-language behavior, expected fixes and exclusions, required
tests, gate policy, conflict group, and priority. Parked failures retain
a reason. Only the behavior is injected as an instruction into query
decomposition and final synthesis; structured fields support routing
and gating. No executable skill code is generated. Superseded entries
retain their history while becoming inactive.

\paragraph{Admission.}
Automatic admission checks targeted improvement, zero protected
regressions, and zero false triggers on no-op, multi-period, and
cross-company negative controls. The run specification also requires
non-decreasing evidence support and factuality; the reference-based
judge does not independently verify every claim against retrieved
evidence. Candidates may additionally be refused or deferred for
insufficient benefit, unstable triggering, or conflicts. High-risk or
unclear candidates have a financial-expert review route; no such expert
was available, so this route defaulted to pass.

\section{Twelve-Round Evolution Record}
\label{app:revision-evolution}

The evolution run uses three disjoint question sets: 300 development questions, a protected suite of
120 questions (90 previously correct and 30 high-risk cases), and a
200-question final monitoring set evaluated once per round. Development
data support diagnosis and targeted validation; protected data are
reserved for gating. The proposal cap is three per round, the
supporting-failure threshold is eight per recorded operational mode,
and the skill-text budget has a soft limit of 500 tokens. No stopping
protocol is specified for this R0--R12 record. This evolution run is
separate from the main benchmark, ablation, and efficiency evaluations.

\begin{table}[t]
\centering
\small
\setlength{\tabcolsep}{3pt}
\begin{tabular}{@{}lrrrrcr@{}}
\toprule
Round & Prop. & Acc. & Dev & Final & E/A & Tok. \\
\midrule
R0  & -- & -- & 90 & 40 & 0/0 & 0 \\
R1  & 3 & 1 & 81 & 37 & 1/1 & 165 \\
R2  & 3 & 1 & 73 & 33 & 2/2 & 255 \\
R3  & 3 & 0 & 73 & 33 & 2/2 & 255 \\
R4  & 3 & 1 & 67 & 34 & 3/3 & 342 \\
R5  & 3 & 1 & 61 & 28 & 4/4 & 421 \\
R6  & 3 & 0 & 61 & 28 & 4/4 & 421 \\
R7  & 3 & 1 & 58 & 27 & 5/4 & 386 \\
R8  & 3 & 0 & 58 & 27 & 5/4 & 386 \\
R9  & 3 & 0 & 58 & 27 & 5/4 & 386 \\
R10 & 2 & 0 & 58 & 27 & 5/4 & 386 \\
R11 & 2 & 0 & 58 & 27 & 5/4 & 386 \\
R12 & 2 & 1 & 55 & 25 & 6/4 & 347 \\
\bottomrule
\end{tabular}
\caption{R0--R12 ledger. Prop./Acc.: proposed/accepted updates;
Dev/Final: non-correct counts out of 300/200, including partial,
incorrect, and failure verdicts. E/A: ever-promoted/active entries.
Tok.: static active skill-text total, not mean selective injection.}
\label{tab:revision-rounds}
\end{table}

\paragraph{Promotion accounting.}
There were 33 proposals, six promotions, and 27 non-promotions,
including refusals, deferrals, and a withdrawal. The two R11 proposals
are counted in the overview without individual identities or
dispositions. Five candidates explicitly failed protected checks:
one at R2, two at R3, one at R4, and one at R9. Their tests recorded
$2+1+2+1+1=7$ item-regression events, which need not involve seven
distinct questions. These candidate tests are distinct from the
zero post-deployment protected regressions reported in every round.

\paragraph{Lifecycle of admitted skills.}
Table~\ref{tab:promoted-lifecycle} follows all six promoted versions from
admission through revision, replacement, and retirement. An active skill
can be revised or have its routing changed; a watch-list entry remains
active until a removal decision. The candidate index below also records
proposals that never entered the active registry.

\begin{table*}[!t]
\centering
\footnotesize
\setlength{\tabcolsep}{4pt}
\renewcommand{\arraystretch}{1.08}
\begin{tabularx}{\textwidth}{@{}>{\raggedright\arraybackslash}p{0.23\textwidth}>{\raggedright\arraybackslash}p{0.19\textwidth}>{\raggedright\arraybackslash}X>{\raggedright\arraybackslash}p{0.13\textwidth}@{}}
\toprule
\textbf{Version / admission} & \textbf{Admission evidence} & \textbf{Revision and maintenance history} & \textbf{R12 status / tokens} \\
\midrule
\nolinkurl{period_entity_alignment_v1}\par R1 &
Period errors $23\rightarrow7$; zero protected regressions. &
Quarterly-scope amendments were deferred at R2 and R4. At R7, v2 covered the earlier behavior and added quarterly scope and entity disambiguation; v1 left the active set after the replacement check. &
Superseded at R7\par 165 tokens \\
\addlinespace[4pt]
\nolinkurl{table_unit_semantics_v1}\par R2 &
Unit errors $17\rightarrow5$; gate passed. &
R5 routing gave the calculation skill priority on calculation queries and retained this skill for unit checking. A broader v2 was rejected as redundant at R6; a template amendment was deferred at R8. &
Active\par 90 tokens \\
\addlinespace[4pt]
\nolinkurl{answer_slot_coverage_v1}\par R4 &
Slot errors $15\rightarrow6$; gate passed. &
The v2 proposal was rejected as redundant at R8. Low activation prompted watch-list review at R10. At R12, the fusion skill covered its fallback behavior; a removal test found zero protected regressions, and v1 was retired. &
Retired at R12\par 87 tokens \\
\addlinespace[4pt]
\nolinkurl{cited_operand_calculation_v1}\par R5 &
Calculation errors $11\rightarrow4$; gate passed. &
R5 priority rules resolved its interaction with unit checking. At R11, over-activation on simple year-on-year queries prompted tighter router triggers; the skill text was retained. &
Active\par 79 tokens \\
\addlinespace[4pt]
\nolinkurl{period_entity_alignment_v2}\par R7 &
Coverage check for replacing v1; zero protected regressions. &
Combined period alignment, quarterly scope, and entity disambiguation. It replaced v1 with a shorter instruction; the separate entity-disambiguation v3 proposal was rejected as redundant. &
Active\par 130 tokens \\
\addlinespace[4pt]
\nolinkurl{evidence_fusion_tiebreak_v1}\par R12 &
Fusion errors $8\rightarrow2$; zero protected regressions. &
Fusion merge v1 failed a protected check at R9; merge v2 was withheld for unstable triggering at R10. This skill was admitted after a diagnosis update and covered the slot-coverage fallback. &
Active\par 48 tokens \\
\bottomrule
\end{tabularx}
\caption{Lifecycle of the six promoted versions in the twelve-round run.
Error counts refer to each candidate's targeted validation.
Token counts are per-version instruction lengths, including inactive
versions. These records are separate from the in-house ablation skills
in Table~\ref{tab:inhouse-skill-records}.}
\label{tab:promoted-lifecycle}
\end{table*}

\paragraph{Interactions between skills.}
Table~\ref{tab:skill-governance} records the six conflict events.
Static checks stopped redundant or contradictory proposals before
admission. Coverage checks supported version replacement, while
activation monitoring led to routing changes for skills already in use.

\begin{table*}[!t]
\centering
\footnotesize
\setlength{\tabcolsep}{4pt}
\renewcommand{\arraystretch}{1.08}
\begin{tabularx}{\textwidth}{@{}>{\raggedright\arraybackslash}p{0.14\textwidth}>{\raggedright\arraybackslash}p{0.36\textwidth}>{\raggedright\arraybackslash}X@{}}
\toprule
\textbf{Round / event} & \textbf{Interaction} & \textbf{Decision and effect} \\
\midrule
R5 / Conflict-C4 &
Calculation instructions overrode unit declarations when both skills were active. &
Give calculation instructions priority on calculation queries, retain their unit-check fields, and use the unit skill for auxiliary checking. \\
\addlinespace[3pt]
R6 / Conflict-C1 &
The proposed unit-semantics v2 duplicated the active v1. &
Reject v2 during static checking, before gate evaluation. \\
\addlinespace[3pt]
R7 / Conflict-C2 &
The alignment v2 trigger overlapped with v1. &
Check coverage and replace v1 with v2, retaining one active alignment version. \\
\addlinespace[3pt]
R8 / Conflict-C3 &
The slot-coverage v2 proposal duplicated the existing behavior. &
Reject the redundant candidate and retain v1. \\
\addlinespace[3pt]
R8 / Conflict-C5 &
Blanket abstention on conflicting evidence contradicted the slot skill's requirement to explain gaps and provide the supported answer. &
Reject the abstention candidate during static checking. \\
\addlinespace[3pt]
R11 / Conflict-C6 &
The calculation skill over-activated on simple year-on-year queries. &
Restrict routing to explicit calculation intent or multi-operand structure; keep the skill text unchanged. \\
\bottomrule
\end{tabularx}
\caption{Conflict handling across the skill lifecycle. Event identifiers
follow the report and use the \texttt{Conflict-} prefix introduced in
Appendix~\ref{app:revision-prompts}.}
\label{tab:skill-governance}
\end{table*}

\paragraph{Skill-function index.}
\label{app:skill-function-index}
Table~\ref{tab:skill-function-index} groups 31 named proposal entries
into 18 functions and lists their recorded outcomes. The entries include
amendments and repeated reviews. Versions share a function; evidence-fusion
and policy-recency variants are grouped by function. The two unnamed R11
entries are outside this index. The descriptions summarize the run record.

\begin{table*}[t]
\centering
\footnotesize
\setlength{\tabcolsep}{4pt}
\renewcommand{\arraystretch}{1.05}
\begin{tabularx}{\textwidth}{@{}>{\raggedright\arraybackslash}p{0.20\textwidth}>{\raggedright\arraybackslash}p{0.34\textwidth}>{\raggedright\arraybackslash}X@{}}
\toprule
Function & Recorded candidates / rounds & Outcome and maintenance \\
\midrule
Period/Entity Alignment &
\nolinkurl{period_entity_alignment_v1} (R1); quarterly amendments (R2, R4); \nolinkurl{period_entity_alignment_v2} (R7) &
v1 promoted; amendments deferred. v2 added quarterly scope and entity disambiguation, superseding v1 at R7; active at R12. \\
\addlinespace[2pt]
Entity Disambiguation &
\nolinkurl{entity_disambiguation_v1/v2/v3} (R4/R5/R7) &
v1 failed a protected check; v2 had insufficient benefit; v3 was rejected as redundant with alignment v2. \\
\addlinespace[2pt]
Evidence Fusion &
\nolinkurl{evidence_fusion_merge_v1/v2} (R9/R10); \nolinkurl{evidence_fusion_tiebreak_v1} (R12) &
v1 caused a protected regression; v2 was withheld for unstable triggering. The revised procedure was promoted at R12 and remained active. \\
\addlinespace[2pt]
Policy Recency &
\nolinkurl{d2_context_priority_v1} (R9, R10); \nolinkurl{d2_policy_recency_v1} (R12) &
Deferred, then declined for insufficient recurring failures; the remaining issue became a known limitation. \\
\addlinespace[2pt]
Table Unit Semantics &
\nolinkurl{table_unit_semantics_v1} (R2), \nolinkurl{table_unit_semantics_v2} (R6); template amendment (R8) &
v1 promoted and active at R12; routing priority adjusted at R5. v2 rejected as redundant; amendment deferred. \\
\addlinespace[2pt]
Answer Slot Coverage &
\nolinkurl{answer_slot_coverage_v1} (R4); \nolinkurl{answer_slot_coverage_v2} (R8) &
v1 promoted; v2 rejected as redundant. v1 entered dormancy review at R10 and was retired at R12 after a protected removal test. \\
\addlinespace[2pt]
Cross-Filing Chronology &
\nolinkurl{cross_filing_date_order_v1} (R6, R7) &
Deferred, then withdrawn after alignment v2 addressed most reviewed cases. \\
\addlinespace[2pt]
Answer Length &
\nolinkurl{answer_length_guard_v1} (R3) &
Declined as an output-style adjustment outside the reliability-repair scope. \\
\addlinespace[2pt]
Cited Calculation &
\nolinkurl{cited_operand_calculation_v1} (R5) &
Promoted and active at R12; routing priority changed at R5 and triggering restricted at R11. \\
\addlinespace[2pt]
Company Aliases &
\nolinkurl{company_alias_table_v1} (R1) &
Declined because the supporting failure count was below threshold. \\
\addlinespace[2pt]
Conservative Abstention &
\nolinkurl{conservative_abstention_v1} (R8) &
Rejected for conflicting with the existing answer-slot behavior. \\
\addlinespace[2pt]
Date Normalization &
\nolinkurl{fiscal_calendar_normalizer_v1} (R1) &
Declined: date formatting did not address incorrect fiscal-period selection. \\
\addlinespace[2pt]
Fiscal Year Assertion &
\nolinkurl{fiscal_period_assert_v1} (R6) &
Rejected for overlap with alignment v1; directed to version revision. \\
\addlinespace[2pt]
Multi-Hop Decomposition &
\nolinkurl{multi_hop_decomposer_v1} (R3) &
Failed protected checks: over-decomposition diverted retrieval on simple queries. \\
\addlinespace[2pt]
Numeric Bounds &
\nolinkurl{numeric_sanity_bound_v1} (R9) &
Declined because financial magnitudes vary across industries, limiting fixed bounds. \\
\addlinespace[2pt]
Numeric Citation &
\nolinkurl{evidence_quote_required_v1} (R3) &
Declined after reduced clarity and a protected regression. \\
\addlinespace[2pt]
Summary Priority &
\nolinkurl{summary_over_narrative_v1} (R5) &
Rejected for overlap with reranking and the risk of worsening summary-over-body preference. \\
\addlinespace[2pt]
Table Row Lock &
\nolinkurl{table_row_lock_v1} (R2) &
Failed protected checks: legitimate queries required aggregation across rows. \\
\bottomrule
\end{tabularx}
\caption{Functional index of the named R1--R12 proposal entries.
Version suffixes separated by slashes denote separate candidates.
Active status refers to the end of R12.}
\label{tab:skill-function-index}
\end{table*}

\paragraph{Final registry.}
Four versions remained active at R12: period/entity alignment v2,
table-unit semantics v1, cited-operand calculation v1, and evidence-fusion
tiebreak v1. Their static instruction lengths totalled
$130+90+79+48=347$ tokens. Alignment v2 replaced a 165-token v1 with
130 tokens at R7. The R12 fusion skill added 48 tokens while slot-coverage
retirement removed 87, reducing the active text from 386 to 347 tokens.
The fusion behavior aligns period/entity/accounting-basis triplets,
merges consistent evidence, and describes inconsistent evidence separately.

\paragraph{Interventions and cost.}
The R9 fusion candidate reduced targeted errors from eight to five
but caused one protected regression. At R10, its revision passed
protected checks but was withheld for unstable diagnosis-dependent
triggering. A diagnostic upgrade preceded the R12 promotion. Routing
also changed: an R5 priority adjustment resolved a unit/calculation
interaction, and an R11 hotfix restricted over-triggering on simple
queries detected through production telemetry. Thus, the trajectory
combines skill, routing, and diagnostic changes under fixed model and
retrieval components. It does not isolate a skill-text effect or
establish long-term production generalization. The Final count
temporarily worsened at R4. At R12 the reported mean prompt is 15,480
tokens; $347/15{,}480\approx2.2\%$ compares static skill text with this
mean, not independently measured runtime injection overhead.

\section{Judge--Human Agreement}
\label{app:revision-human}

Two annotators from the author group independently judged a mixed
sample of 150 answers against gold answers and key points. Both had
financial backgrounds, but neither was a financial domain expert;
no external annotator participated. Table~\ref{tab:revision-human}
includes the primary Qwen3-Max judge and an additional GPT-5.5 check.

\begin{table}[t]
\centering
\small
\setlength{\tabcolsep}{3pt}
\begin{tabular}{@{}lcc@{}}
\toprule
Pair & Four-way & Correct / non-correct \\
 & \% / $\kappa$ & \% / $\kappa$ \\
\midrule
A--B & 82.0 / .687 & 92.7 / .853 \\
Qwen3-Max--A & 90.7 / .820 & 91.3 / .827 \\
Qwen3-Max--B & 81.3 / .675 & 92.0 / .840 \\
GPT-5.5--A & 78.0 / .639 & 95.3 / .907 \\
GPT-5.5--B & 81.3 / .703 & 94.7 / .893 \\
\bottomrule
\end{tabular}
\caption{Agreement on the same 150-answer sample.}
\label{tab:revision-human}
\end{table}

The annotators disagreed on 27 four-way and 11 binary decisions.
Human--human binary $\kappa$ has a 95\% bootstrap interval of
$[0.760,0.933]$. Judge--human binary estimates lie within this
interval; this descriptive comparison is not an equivalence test
or evidence of expert-level financial assessment.

\section{Adaptations of Other Skill-Evolution Methods}
\label{app:skill-adapters}

We adapted AutoSkill~\citep{yang2026autoskill} and
SkillOpt~\citep{yang2026skillopt} to the filing-QA substrate, using
DeepSeek-V3.2 for generation and Qwen3-Max with the same evaluation
rubric. The adapters converted input/output schemas, trajectories,
and scores, and inserted skill text into prompts. They left retrieval,
the agent workflow, and the evaluator unchanged, with no method-specific
tuning.

\paragraph{AutoSkill interface.}
For each question, AutoSkill retrieved relevant skills from its own bank
and added them to the answer-generation prompt. After scoring, the
question, answer, and judge feedback were returned to AutoSkill to
update its bank online. This evaluation therefore used a changing bank,
unlike the frozen registry used for Full \evolvename{}.

\paragraph{SkillOpt interface.}
The substrate served as SkillOpt's environment and rollout interface.
SkillOpt revised one Markdown skill using fixed training and validation
sets. After optimization, the best skill was frozen and injected only
into final synthesis for evaluation. Its optimization loop completed
even when no candidate update was accepted.

\begin{table}[!b]
\centering
\small
\setlength{\tabcolsep}{2pt}
\begin{tabular}{@{}lrrrrrr@{}}
\toprule
Method & W-Corr. & Info. & Reas. & Fact. & Clar. & Depth \\
\midrule
\multicolumn{7}{@{}l}{\textbf{Zeekr}} \\
Naive RAG & 2.67 & 2.47 & 2.96 & 3.56 & 4.37 & 2.50 \\
AutoSkill & 2.97 & 2.73 & 2.27 & 3.00 & 4.17 & 2.53 \\
SkillOpt & 3.40 & 3.63 & 2.60 & 3.10 & 4.13 & 2.80 \\
Full \evolvename{} & 4.52 & 4.56 & 4.32 & 4.74 & 4.85 & 4.26 \\
\midrule
\multicolumn{7}{@{}l}{\textbf{Lotus}} \\
Naive RAG & 2.93 & 2.98 & 3.45 & 3.68 & 4.50 & 3.05 \\
AutoSkill & 1.33 & 1.57 & 1.43 & 2.03 & 4.73 & 1.37 \\
SkillOpt & 2.63 & 2.70 & 1.63 & 3.00 & 3.70 & 1.90 \\
Full \evolvename{} & 3.74 & 3.91 & 3.83 & 4.20 & 4.50 & 3.80 \\
\midrule
\multicolumn{7}{@{}l}{\textbf{FinanceBench}} \\
Naive RAG & 2.21 & 2.36 & 2.52 & 2.76 & 4.54 & 2.40 \\
AutoSkill & 2.53 & 2.17 & 2.33 & 2.97 & 4.37 & 2.30 \\
SkillOpt & 2.60 & 3.40 & 3.13 & 2.63 & 4.87 & 2.90 \\
Full \evolvename{} & 3.49 & 3.39 & 3.49 & 3.78 & 4.32 & 3.35 \\
\midrule
\multicolumn{7}{@{}l}{\textbf{FinDER}} \\
Naive RAG & 1.55 & 1.52 & 1.65 & 2.35 & 4.14 & 1.48 \\
AutoSkill & 1.17 & 1.23 & 1.20 & 2.20 & 4.33 & 1.23 \\
SkillOpt & 2.00 & 2.53 & 1.70 & 2.20 & 3.43 & 2.67 \\
Full \evolvename{} & 2.88 & 2.85 & 3.30 & 3.83 & 3.99 & 2.83 \\
\bottomrule
\end{tabular}
\caption{Filing-QA results for the AutoSkill and SkillOpt adapters,
with Naive RAG as the no-skill reference. W-Corr. uses the verdict-score
mapping in Appendix~\ref{app:eval-protocol}; the remaining columns are
the five judge-rated quality dimensions.}
\label{tab:skill-adapter-results}
\end{table}

\paragraph{Benchmark results.}
Table~\ref{tab:skill-adapter-results} reports four benchmarks.
SkillOpt improved W-Corr. over Naive RAG on three, and AutoSkill on two.
On Lotus and FinDER, AutoSkill had the highest clarity but the lowest
correctness.
Full \evolvename{} had the highest correctness on all four.
These results describe the filing-QA adapters with their different update
and injection interfaces; they do not isolate the effect of a gate or
provide a comparison under matched evolution budgets.

\paragraph{Earlier interface tests.}
Separate pilot runs examined whether the methods could extract and
revise skills from filing-QA feedback.
With raw question--answer pairs, AutoSkill produced no skills over
eleven rounds. Wrapping judge feedback as policy corrections enabled
skill extraction: from R00 to R06, its bank grew from zero to four
skills. The injected context reached 36,214 characters, and the answer
prompt grew from 6,105 to 11,551 tokens. Correct counts on the target,
protected, and monitoring splits remained unchanged across these rounds.

SkillOpt was tested at four scales, the largest with 90 training,
22 validation, and 100 test items. The reported optimizer counters were
\texttt{accept=0}, \texttt{reject=0}, and \texttt{skip=6}.
Reflection identified failures in the larger runs but judged them
unsuitable for repair through the skill document, often because
evidence was insufficient. A separate epoch-boundary update path
was enabled with \texttt{use\_slow\_update=true} and
\texttt{slow\_update\_gate\_with\_selection=false}; this path bypassed
validation. Test hard accuracy changed from 0.680 to 0.670 on 100 items,
a net decrease of one correct answer. This pilot used hard accuracy,
separately from the W-Corr. results in Table~\ref{tab:skill-adapter-results}.

\section{Evolution on Public Benchmarks}
\label{app:public-evolution}

We ran six-round evolution pilots on FinanceBench and SECQUE.
Table~\ref{tab:public-evolution} gives the development, protected, and
monitoring allocations and summarizes the runs. Promotion followed
Equation~\ref{eq:gate}. Table~\ref{tab:public-evolution-rounds} reports
each monitoring observation, including the brief drop in SECQUE W-Corr.
at R3. These pilots are separate from the twelve-round study in
Appendix~\ref{app:revision-evolution}.

\begin{table}[!b]
\centering
\small
\setlength{\tabcolsep}{3pt}
\begin{tabularx}{\columnwidth}{@{}Xrr@{}}
\toprule
 & FinanceBench & SECQUE \\
\midrule
Total questions & 150 & 100 \\
Development & 60 & 40 \\
Protected & 30 & 20 \\
Monitoring & 60 & 40 \\
Evolution rounds & 6 & 6 \\
\midrule
Proposed candidates & 15 & 9 \\
Promotions & 3 & 2 \\
Active skills at R6 & 2 & 2 \\
Protected-gate refusals & 2 & 1 \\
Candidate regression events & 3 & 1 \\
Deployed regressions & 0 & 0 \\
Skill-text tokens at R6 & 214 & 163 \\
Reported Target Val.\ $\Delta$ & +0.44 & +0.38 \\
\bottomrule
\end{tabularx}
\caption{Six-round public-benchmark evolution pilots. Candidate regression
events occur during validation; deployed regressions concern the admitted
registry. Target Val.\ $\Delta$ is the reported targeted-validation gain.
Tokens count active skill text.}
\label{tab:public-evolution}
\end{table}

\paragraph{FinanceBench.}
At R2, we promoted \nolinkurl{fiscal_period_anchor_v1} to address
recurring answers that identified the right metric but used the wrong
fiscal year. At R3, \nolinkurl{segment_total_guard_v1} entered the
registry. At R4, \nolinkurl{derived_ratio_check_v1} added a check that
recomputed the denominator when draft percentages conflicted with cited
amounts. By R6, the ratio-check skill had absorbed the segment-total
behavior; the older skill passed a removal test and was retired.

Two other candidates were refused at R3 and R5 after causing three
item-level protected regression events in total. The deployed registry
recorded zero protected regressions throughout the run.

\paragraph{SECQUE.}
At R2, \nolinkurl{filing_section_scope_v1} restricted answers to the
section and entity requested by the question. At R5,
\nolinkurl{numeric_unit_guard_v1} was promoted. This pilot independently
produced a skill with the same name and numeric-unit function as the
in-house skill listed in Table~\ref{tab:inhouse-skill-records}.
A separate candidate was refused at R4 after one protected regression
event.

\paragraph{Registry comparison.}
The in-house registry scored above the locally evolved registry on both
monitoring splits (Table~\ref{tab:public-registry-comparison}). On
FinanceBench, their Correct counts were equal; the gap came from the
non-correct verdicts. The in-house scores differ from
Table~\ref{tab:per_benchmark_results} because the evaluations here use
monitoring subsets.

\begin{table}[!t]
\centering
\small
\setlength{\tabcolsep}{3pt}
\begin{tabular*}{\columnwidth}{@{}l@{\extracolsep{\fill}}ccc@{}}
\toprule
Registry & C/P/I/F & Non-correct & W-Corr. \\
\midrule
\multicolumn{4}{@{}l}{\textbf{FinanceBench (60 monitoring questions)}} \\
No skills & 37/0/2/21 & 38.3\% & 3.12 \\
Local R6 & 42/0/1/17 & 30.0\% & 3.52 \\
In-house & 42/1/2/15 & 30.0\% & 3.58 \\
\midrule
\multicolumn{4}{@{}l}{\textbf{SECQUE (40 monitoring questions)}} \\
No skills & 17/2/0/21 & 57.5\% & 2.28 \\
Local R6 & 20/0/1/19 & 50.0\% & 2.53 \\
In-house & 21/0/0/19 & 47.5\% & 2.63 \\
\bottomrule
\end{tabular*}
\caption{Registry comparison on common monitoring questions under
one frozen judge. C/P/I/F denotes Correct/Partial/Incorrect/Failure
counts. W-Corr. is $(5C+3P+I)/N$, including Failure as zero.
No skills is the R0 condition; Local R6 is the registry evolved on the
respective benchmark. In-house is the registry used for Full in
Table~\ref{tab:per_benchmark_results}, re-evaluated on these subsets.}
\label{tab:public-registry-comparison}
\end{table}

\begin{table}[!b]
\centering
\small
\setlength{\tabcolsep}{2pt}
\begin{tabular*}{\columnwidth}{@{}l@{\extracolsep{\fill}}rrrr@{}}
\toprule
 & \multicolumn{2}{c}{FinanceBench ($n=60$)} & \multicolumn{2}{c}{SECQUE ($n=40$)} \\
\cmidrule(lr){2-3}\cmidrule(l){4-5}
Round & NC (\%) & W-Corr. & NC (\%) & W-Corr. \\
\midrule
R0 & 23 (38.3) & 3.12 & 23 (57.5) & 2.28 \\
R1 & 22 (36.7) & 3.18 & 22 (55.0) & 2.33 \\
R2 & 21 (35.0) & 3.27 & 21 (52.5) & 2.45 \\
R3 & 20 (33.3) & 3.35 & 21 (52.5) & 2.40 \\
R4 & 19 (31.7) & 3.43 & 20 (50.0) & 2.50 \\
R5 & 19 (31.7) & 3.45 & 20 (50.0) & 2.53 \\
R6 & 18 (30.0) & 3.52 & 20 (50.0) & 2.53 \\
\bottomrule
\end{tabular*}
\caption{Monitoring trajectories over six update rounds. NC gives
non-correct counts (percentages); both metrics use the same per-question
verdicts. R0 is the no-skill baseline.}
\label{tab:public-evolution-rounds}
\vspace{-0.3cm}
\end{table}

\end{document}